\documentclass{article} %
\usepackage{iclr2026_preprint,times}

\usepackage{amsmath,amsfonts,bm}

\def\eqref#1{equation~\ref{#1}}

\def\1{\bm{1}}

\DeclareMathAlphabet{\mathsfit}{\encodingdefault}{\sfdefault}{m}{sl}
\SetMathAlphabet{\mathsfit}{bold}{\encodingdefault}{\sfdefault}{bx}{n}

\usepackage{hyperref}
\usepackage{url}

\usepackage[utf8]{inputenc} %
\usepackage[T1]{fontenc}    %
\usepackage{hyperref}       %
\usepackage{url}            %
\usepackage{booktabs}       %
\usepackage{amsfonts}       %
\usepackage{nicefrac}       %
\usepackage{microtype}      %
\usepackage{xcolor}
\usepackage{makecell}%
\usepackage{graphicx}
\usepackage{tcolorbox}
\usepackage{enumitem}
\usepackage{amsmath}
\usepackage{wrapfig}
\usepackage{tikz}
\usetikzlibrary{decorations.pathreplacing,calligraphy}
\usepackage{pgfplots}
\usepackage{pgfplotstable}
\usepackage{xcolor}
\usepgfplotslibrary{polar}
\usepackage{float}
\usepackage{etoc}

\usepackage{longtable}
\usepackage{booktabs}
\usepackage{multirow}
\usepackage{makecell}
\usepackage{caption}

\usepackage{booktabs,multirow,array}

\definecolor{defcolor}{HTML}{f9f9f4} %
\definecolor{framecolor}{HTML}{d3d3d3} %
\definecolor{bordercolor}{HTML}{a6a6a6} %

\definecolor{defcolor}{HTML}{EFF6FC}    %
\definecolor{framecolor}{HTML}{A9CCE3}  %
\definecolor{bordercolor}{HTML}{2980B9} %

\definecolor{defcolor}{HTML}{f5f8fa} %
\definecolor{framecolor}{HTML}{c7d5df} %
\definecolor{bordercolor}{HTML}{5a7d95} %

\tcbset{
  takeawaybox/.style={
    colback=white,
    colframe=finalblue,
    boxrule=1pt,
    arc=1pt,
    left=6pt,
    right=6pt,
    top=6pt,
    bottom=6pt,
    fonttitle=\bfseries,
  }
}

\title{Mapping Post-Training Forgetting in Language Models at Scale}

\author{%
\hspace{0.5cm}\textbf{Jackson Harmon} \qquad  
\textbf{Andreas Hochlehnert}\qquad
\textbf{Matthias Bethge}\thanks{Equal Supervision} \qquad 
\textbf{Ameya Prabhu}\textsuperscript{*} \\ \vspace{0.5cm}
\centerline{Tübingen AI Center, University of Tübingen}
}

\newcommand{\invisfootnote}[1]{%
  \begingroup
  \renewcommand\thefootnote{}%
  \footnotetext{#1}%
  \endgroup
}

\iclrfinalcopy %

\newtcolorbox{takeawaybox}{
  colback=orange!10,    %
  colframe=orange!70,   %
  boxrule=0.5pt,
  arc=2pt,
  left=6pt,
  right=6pt,
  top=4pt,
  bottom=4pt,
  title=Takeaway
}

\newtcolorbox{keyfindings}{
  colback=blue!5!white,
  colframe=blue!65!black,
  title=Key Findings,
  boxrule=0.5pt,
  arc=2pt,
  left=6pt,
  right=6pt,
  top=4pt,
  bottom=4pt,
}

\newcommand{\takeaway}[1]{%
  \begin{takeawaybox}
  #1
  \end{takeawaybox}
}

\pgfplotsset{compat=1.18}

\begin{document}

\maketitle
\lhead{Preprint}  %

\invisfootnote{Code, data, and interactive plots are available at \url{https://post-forget.github.io/}}

\vspace{-0.35cm}
\begin{abstract}
Scaled post‑training now drives many of the largest capability gains in language models (LMs), yet its effect on pretrained knowledge remains poorly understood. Not all forgetting is equal: Forgetting one fact (e.g., a U.S. president or an API call) does not “average out” by recalling another. Hence, we propose a sample-wise paradigm to measure what is forgotten and when backward transfer occurs. Our metric counts 1→0 transitions (correct before post‑training, incorrect after) to quantify forgetting and 0→1 transitions to quantify backward transfer. Traditional task averages conflate these effects and obscure large changes. For multiple‑choice benchmarks, we add chance‑adjusted variants that subtract the expected contribution of random guessing from pre‑ and post‑training accuracies. We apply this framework across post‑training stages, model sizes, and data scales. Our large‑scale analysis shows that: (1) Domain-continual pretraining induces moderate forgetting with low-to-moderate backward transfer; (2) RL/SFT post-training applied to base models and Instruction tuning yields moderate-to-large backward transfer on math and logic with overall low-to-moderate forgetting; (3) Applying RL/SFT to instruction‑tuned models is sensitive on data scale: at small scales, both forgetting and backward transfer are small; at larger scales, effects are mixed and warrant further study with better controls; (4) Model merging does not reliably mitigate forgetting. Overall, our framework offers a practical yardstick for mapping how post‑training alters pretrained knowledge at scale -- enabling progress towards generally capable AI systems.
\end{abstract}

\vspace{-0.35cm}
\section{Introduction}
\vspace{-0.15cm}

Scaling post-training has become the dominant driver of capability gains in modern language models (LMs) \citep{jaech2024openai}. Practitioners now iterate through multi-step post-training pipelines often at data scales that rival early pretraining~\citep{tie2025surveyposttraininglargelanguage}. The implicit bet is that each step in the pipeline accumulates new capabilities, with dramatic improvements in areas like coding, math, tool use and safety, without sacrificing the broad world knowledge. In contrast, it is considered common knowledge in continual learning that this sequential training would lead to catastrophic forgetting (see Table~\ref{tab:forgetlit}). We test this assumption: as we scale post-training, do we erode the very breadth of world knowledge that pretraining painstakingly compresses into the weights? If the implicit assumption does not hold, we risk trading generalist competence for narrow specialization, undermining progress toward generally capable models. 

Measuring forgetting in modern post-training pipelines is tricky. Classical evaluations compare aggregate test accuracy before and after training \citep{luo2025empiricalstudycatastrophicforgetting}, implicitly treating a benchmark as a single task with fungible i.i.d. samples (e.g., classifying images of cats). Pretrained knowledge violates this assumption. Knowing one U.S. president does not compensate for forgetting another; recalling a  NumPy broadcasting rule does not offset losing a specific cloud-API syntax. In short, knowledge samples are not fungible: Each carries unique value for quantifying pretraining knowledge. Aggregation can hide substantial losses. Hence, we measure forgetting and backward transfer in a sample-wise manner, rather than at the task level as proposed by \cite{lopez2017gradient}. 

Specifically, we define \emph{forgetting} as items that are answered correctly before a post-training stage but incorrectly afterward (the \(1\!\rightarrow\!0\) transitions), and \emph{backward transfer} as items that are answered incorrectly before but correctly after post-training (the \(0\!\rightarrow\!1\) transitions). A further complication is that most knowledge-intensive LLM evaluation benchmarks are multiple-choice. Random guessing inflates accuracy and can create illusory transitions: an apparent “\(1\!\rightarrow\!0\)” may simply be a lucky guess that later becomes an incorrect answer, even when the underlying knowledge did not change; likewise for \(0\!\rightarrow\!1\) transitions. When the answer is only among few options (e.g., 4), performance by random guessing can account for a substantial share of observed transitions, distorting both level and trend estimates of forgetting. Thus a principled metric should (i) resolve outcomes at the \emph{item} level and (ii) explicitly correct for chance.

We introduce chance-adjusted metrics for forgetting ($\mathrm{F}_\text{true}$) and backward transfer ($\mathrm{BT}_\text{true}$), which correct for transitions expected under random choice. They do not need logits or repeated sampling, measurable using the number of choices in benchmark and marginal accuracy of the model pre- and post- training, making them practical at scale. Intuitively, chance-adjusted forgetting asks: among items the model genuinely knew before, what fraction became wrong beyond chance? Conversely, chance-adjusted backward asks: among items the model genuinely did not correctly solve, what fraction became correct beyond chance?

Our primary contribution is a large-scale study measuring forgetting caused by post-training across post-training pipelines. By evaluating the models on the same set of samples before and after each stage, we obtain a map of what was retained, what was forgotten, and where losses concentrate. We seek to answer three questions: (i) Where in the pipeline is forgetting most pronounced (e.g., instruction tuning vs. reasoning-focused training)?, (ii) What kinds of pretraining knowledge are most affected (culture vs. logic)?, and (iii) How much knowledge is forgotten or re-elicited? We have the following key findings:

\begin{keyfindings}
\begin{itemize}[leftmargin=1em]
    \item \textbf{Domain-Continual Pretraining} induces low to moderate forgetting across most categories; backward transfer is limited. Forgetting effects marginally decrease with increasing model scale.
    \item \textbf{Instruction-Tuning and SFT/RL from base models} yield low to moderate forgetting, with spikes in the Culture and Knowledge categories, but moderate to high (for SFT/RL from Base) backward‑transfer gains in the Math and Logic categories across model families; Forgetting and backtransfer decrease as parameters increase. Reasoning training yields similar forgetting and larger backward transfer than instruction tuning.
    \item \textbf{SFT/RL Reasoning Post-Training from instruct models} have data-scale dependent behaviour: For the low‑data regime, it yields low forgetting and backward transfer. For the high-data regime, no dominant factor robustly described the forgetting and backward transfer dynamics.
    \item \textbf{Model Merging} does not reliably mitigate forgetting across post-training pipelines (yet).
\end{itemize}

\end{keyfindings}

\begin{longtable}{p{0.08\textwidth} p{0.38\textwidth} p{0.05\textwidth} p{0.35\textwidth}}
\caption{\textbf{Catastrophic forgetting literature across LLM post-training stages.} Continual learning literature indicates extensive forgetting across the post-training pipeline. However, we find far less forgetting when testing widely used post-training pipelines, indicating an important gap existing between continual learning setups and how people post-train language models.}
\label{tab:forgetlit} \\
\toprule
\textbf{Stage} & \textbf{Name} & \textbf{Level} & \textbf{Summary} \\
\midrule
\endfirsthead

\multicolumn{4}{l}{\textit{(Continued from previous page)}} \\
\toprule
\textbf{Stage} & \textbf{Name} & \textbf{Level} & \textbf{Summary} \\
\midrule
\endhead

\midrule
\multicolumn{4}{r}{\textit{(Continued on next page)}} \\
\endfoot

\bottomrule
\endlastfoot

\multirow{2}{*}{\makecell{CPT \\(§\ref{subsec:da})}}
& Investigating Continual Pretraining in LLMs: Insights and Implications \citep{yildiz2024investigating}
& Med
& Most models show continual improvement; only Llama-2 models degrade. \\
& Examining Forgetting in Continual Pre-training of Aligned LLMs \citep{li2024examining}
& High
& Continual pre-training degrades capabilities, alignment and alters output behavior. \\
\midrule
SFT/DPO (§\ref{subsec:it})
& Mitigating Forgetting in LLM Supervised Fine-Tuning and Preference Learning \citep{fernando2024mitigating}
& Low
& Combining SFT and DPO sequentially leads to forgetting and a poor balance between goals ($\sim2\%$ on MMLU). \\
\midrule
\multirow{4}{*}{\makecell{SFT\\(§\ref{subsec:sft_rl})}}
& Interpretable Catastrophic Forgetting of LLM Fine-tuning via Instruction Vector \citep{jiang2024refine}
& High
& Fine-tuning on TRACE shows declines primarily from lost instruction-following ability. \\
& An Empirical Study of Catastrophic Forgetting in LLMs During Continual Fine-tuning \citep{luo2025empiricalstudycatastrophicforgetting}
& High
& Forgetting of domain knowledge, reasoning intensifies as model scale increases ($\sim10\%$ MMLU drop). \\
& Catastrophic Forgetting in LLMs: A Comparative Analysis Across Language Tasks \citep{haque2025catastrophic}
& High
& Severity varies by architecture and pre-training quality; some models degrade sharply while others barely change. \\
& Mitigating Catastrophic Forgetting in LLMs with Self-Synthesized Rehearsal \citep{huang2024mitigating}
& High
& Sequential fine-tuning causes major forgetting; synthetic rehearsal mitigates it. \\
\midrule
RL (§\ref{subsec:it})
& Mitigating the Alignment Tax of RLHF \citep{lin2024mitigatingalignmenttaxrlhf}
& Med
& RLHF induces forgetting (“alignment tax”); model averaging reduces it. \\
\midrule
\multirow{2}{*}{\makecell{SFT/RL\\(§\ref{subsec:it})}}
& Understanding Catastrophic Forgetting in LLMs via Implicit Inference \citep{kotha2024understanding}
& High
& Fine-tuning skews the model’s implicit task inference rather than erasing capabilities. \\
& Temporal Sampling for Forgotten Reasoning in LLMs \citep{li2025temporal}
& High
& Fine-tuned LLMs often forget solutions they previously generated (“temporal forgetting”) across sizes and methods (SFT, GRPO). \\
\end{longtable}
\vspace{0.2cm}

\section{Measuring Samplewise Forgetting and Backward Transfer}
\label{sec:metric}

\begin{wrapfigure}[15]{r}{6.0cm} %
  \centering
  \vspace{-0.15cm}
  \begin{tikzpicture}[x=2cm, y=2cm, font=\footnotesize] %
    \draw[thick] (0,0) rectangle (2,2);
    \draw (1,0) -- (1,2);
    \draw (0,1) -- (2,1);

    \node at (1,-0.45) {\large $a^{\mathrm{pre}}$};
    \node[rotate=90] at (-0.45,1) {\large $a^{\mathrm{post}}$};

    \node at (0.5,-0.2) {0};
    \node at (1.5,-0.2) {1};
    \node at (-0.2,0.5) {0};
    \node at (-0.2,1.5) {1};

    \tikzset{cell/.style={align=center, text width=1.6cm, inner sep=1.5pt}}

    \node[cell] at (0.5,1.5) {Backward\\Transfer};
    \node[cell] at (1.5,0.5) {Forgetting};
    \node[cell] at (1.5,1.5) {Retention};
    \node[cell] at (0.5,0.5) {Non-\\acquisition};
  \end{tikzpicture}
  \caption{Each sample is assigned to one of four quadrants by correctness before and after.}
  \label{fig:tikzquad}
\end{wrapfigure}
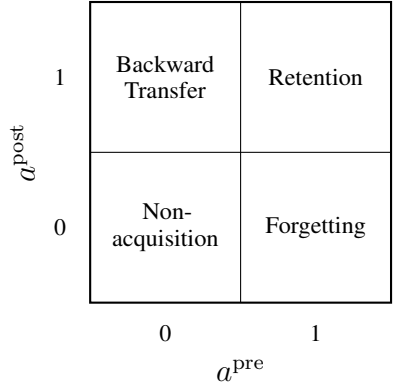

Consider an evaluation set of \(N\) multiple‑choice questions with \(k\) options. For each sample \(i\), let \(a^{\text{pre}}_i,a^{\text{post}}_i\in\{0,1\}\) indicate correctness before and after post‑training. As illustrated in Fig.~\ref{fig:tikzquad}, each sample falls into one of four quadrants based on effect by training on new task: 

(i) Retention preserves knowledge \((1\!\to\!1)\),\\ 
(ii) Backward Transfer improves performance \((0\!\to\!1)\),\\
(iii) Forgetting reduces performance \((1\!\to\!0)\), and \\
(iv) non‑acquisition has no effect \((0\!\to\!0)\).

We define sample‑wise \emph{forgetting} and \emph{backward transfer} as the proportions of \(1\!\to\!0\) and \(0\!\to\!1\) flips, respectively:
\begin{align*}
\mathrm{F}  &= \frac{1}{N}\sum_{i=1}^N \mathbf{1}\{a^{\text{pre}}_i=1\wedge a^{\text{post}}_i=0\}\\
\mathrm{BT} &= \frac{1}{N}\sum_{i=1}^N \mathbf{1}\{a^{\text{pre}}_i=0\wedge a^{\text{post}}_i=1\}
\label{eq:fobs}
\end{align*}

These intuitive metrics confound genuine knowledge change with label flips caused by guessing, especially when \(k\) is small. For example, two independent random binary classifiers (\(k{=}2\)) yield \(\mathrm{F}=0.25\) because \(0.5\times 0.5=0.25\).

\paragraph{A chance baseline for flips.} \begin{wrapfigure}{r}{0.3\textwidth}
\vspace{-20pt}
\centering
\begin{tikzpicture}[font=\large]
    \coordinate (A) at (0,0);
    \coordinate (B) at (4,1);
    \draw[thick] (A) rectangle (B);

    \coordinate (Split1) at (2.5,0); %
    \coordinate (Split2) at (3,0); %
    \draw[thick] (Split1) -- (Split1 |- B);
    \draw[thick] (Split2) -- (Split2 |- B);

    \node at (1.4, 0.5) {$\bar{a}_\text{true}$};
    \node at (2.75, 0.5) {$x$};

    \draw [decorate,decoration={brace,amplitude=5pt,raise=1pt},thick] (0,1.1) -- (3,1.1) 
          node [black,midway,yshift=0.4cm] {$\bar{a}$};

    \draw [decorate,decoration={brace,amplitude=5pt,raise=1pt},thick] (3,1.1) -- (4,1.1) 
          node [black,midway,yshift=0.4cm] {$1-\bar{a}$};

\end{tikzpicture}
\caption{Accuracy \(\bar a\) decomposes into true knowledge \(\bar a_{\text{true}}\) and lucky guesses \(x\).}
\label{fig:actual_guess}
\vspace{-10pt}
\end{wrapfigure}
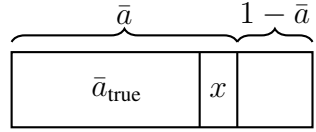

We assume a simple response model: on each item the model either \emph{knows} the answer or \emph{guesses} uniformly among the \(k\) choices. Let \(\bar a\) be mean accuracy on a set. Then \(\bar a=\bar a_{\text{true}}+x\), where \(x\) is the fraction correct by chance. Since an incorrect guess occurs with probability \((k-1)/k\), 
\[
\frac{1-\bar a}{\,x + (1-\bar a)\,}=\frac{k-1}{k}
\quad\Longrightarrow\quad
x=\frac{1-\bar a}{k-1}.
\]
A \(1\!\to\!0\) flip due purely to chance requires (i) a pre‑training correct guess and (ii) a post‑training error (converse for backward transfer). Assuming independence between pre‑ and post‑training guessing events,
\[
\mathrm{F}_{\text{chance}}
=\underbrace{\frac{1-\bar a^{\text{pre}}}{k-1}}_{\text{correct by chance (pre)}}\cdot
\underbrace{(1-\bar a^{\text{post}})}_{\text{incorrect (post)}},
\qquad
\mathrm{BT}_{\text{chance}}
=\underbrace{(1-\bar a^{\text{pre}})}_{\text{incorrect (pre)}}\cdot
\underbrace{\frac{1-\bar a^{\text{post}}}{k-1}}_{\text{correct by chance (post)}}.
\]
These metrics depend only on aggregate accuracies and \(k\); they require no logits or heavy computation.

\paragraph{Chance‑adjusted forgetting and backward transfer.}
To isolate knowledge change beyond chance, subtract the baselines and clip at zero:
\[
\mathrm{F}_{\text{true}}=\max\!\bigl(\mathrm{F}-\mathrm{F}_{\text{chance}},\,0\bigr),\qquad
\mathrm{BT}_{\text{true}}=\max\!\bigl(\mathrm{BT}-\mathrm{BT}_{\text{chance}},\,0\bigr).
\]
For example, if accuracy drops from 80\% to 70\% on a 4-option MCQ test, raw forgetting is 10\%, but chance-adjusted forgetting is only about 6\% -- showing how the correction removes the effect of lucky guesses. Clipping ensures the metric remains valid even if models perform below chance.

\paragraph{Ceilings: how much could a model forget or improve?}
Observed forgetting can be small simply because little was truly correct to begin with. The \emph{maximum possible} forgetting equals the fraction truly correct before post‑training:
\[
\mathrm{F}_{\max}=\bar a^{\text{pre}}_{\text{true}}
=\bar a^{\text{pre}}-x^{\text{pre}}
=\max\!\left(\frac{k\,\bar a^{\text{pre}}-1}{k-1},\,0\right).
\]
Similarly, the \emph{maximum possible} backward transfer equals the fraction truly correct after post‑training:
\[
\mathrm{BT}_{\max}=\bar a^{\text{post}}_{\text{true}}
=\bar a^{\text{post}}-x^{\text{post}}
=\max\!\left(\frac{k\,\bar a^{\text{post}}-1}{k-1},\,0\right).
\]
By construction \(\mathrm{F}_{\text{true}}\le \mathrm{F}_{\max}\) and \(\mathrm{BT}_{\text{true}}\le \mathrm{BT}_{\max}\). Reporting the adjusted metrics alongside these ceilings separates true knowledge loss/acquisition from chance and contextualizes headroom for degradation or improvement.

\paragraph{Assumptions and scope.} The correction uses two assumptions: (i) when the model does not know an answer, it guesses uniformly at random; and (ii) pre‑ and post‑training guessing events are independent. These assumptions allow dataset‑level adjustments from pre‑ and post‑training accuracies alone. Note that \(\mathrm{F}_{\text{true}}\) could quantify failure to elicit previously accessible knowledge and need not imply that the model has lost/unlearned the underlying information. Likewise, changes in \(\mathrm{BT}_{\text{true}}\) often reflect improved elicitation rather than newly acquired knowledge.

\vspace{-0.15cm}
\section{When, What \& How Much is Pretraining Knowledge Forgotten?}
\vspace{-0.15cm}
\label{sec:results}

In this section, we ask three questions: %

\begin{enumerate}
    \item \emph{When is pretraining knowledge forgotten?}\\Our analysis spans four widely used continual‑training regimes: (i) domain‑continual  training (\S\ref{subsec:da}), (ii) instruction tuning (\S\ref{subsec:it}), (iii) light SFT/RL on reasoning traces, and (iv) large‑scale SFT/RL for reasoning (\S\ref{subsec:sft_rl}). In total, we evaluate almost 30 model–training combinations chosen to reflect common practice results, providing broad coverage of how contemporary LLMs are post‑trained in the wild. Each post‑trained model is compared with its initial checkpoint (details in the Appendix).
    \item \emph{What pretraining knowledge is forgotten?}\\We evaluate each model on 12 public benchmarks, collectively subdivided into close to a 100 total subdomains. To summarize systematic patterns, we cluster sub‑benchmarks into nine semantically coherent groups that exhibit similar forgetting trends (e.g., common sense, culture, deduction, language/communication, liberal arts, science/tech). These clusters provide a better map of which pretraining knowledge areas are most affected by a given post‑training recipe.
    \item \emph{How much pretraining knowledge is forgotten?}\\Unless stated otherwise, chance‑adjusted metrics for forgetting ($\mathrm{F}_{\text{true}}$) and backward transfer ($\mathrm{BT}_{\text{true}}$) are used to quantify the severity.
\end{enumerate}

\textbf{Experimental setup.} We standardize settings across models for fair comparison. All experiments use the \texttt{LightEval} framework \citep{lighteval} and log per‑sample accuracy. We apply a zero‑shot chain‑of‑thought prompt to all models and require answers in a fixed MCQ format (see Appendix); base models receive a few‑shot prompt solely to teach the format. When available\footnote{This budget was sufficient in practice, we never required more tokens.}, we add chat‑specific templates to be in line with best practices. We cap sequence length at 32K tokens, except for Qwen2.5‑7B‑Math and Qwen2.5‑7B‑Math‑Instruct~\cite{qwen2.5}, which are limited to 4K\footnote{Because base models sometimes continue into subsequent questions, we set explicit stop sequences to end generation once a prediction is produced.}. Decoding uses temperature $0.6$ with nucleus sampling (\texttt{top\_p}) of $0.95$. We provide additional details in the Appendix. To facilitate reproducibility and further inquiry, we will release per‑sample logs for every sub‑benchmark alongside code. 

We now showcase our results in the subsections below.

\vspace{-0.15cm}
\subsection{Subarea 1: Domain‑Continual Pretraining}
\label{subsec:da}
\vspace{-0.1cm}

\begin{figure}[htbp]
  \centering
  \vspace{-0.3cm}
  \includegraphics[width=0.95\textwidth,keepaspectratio]{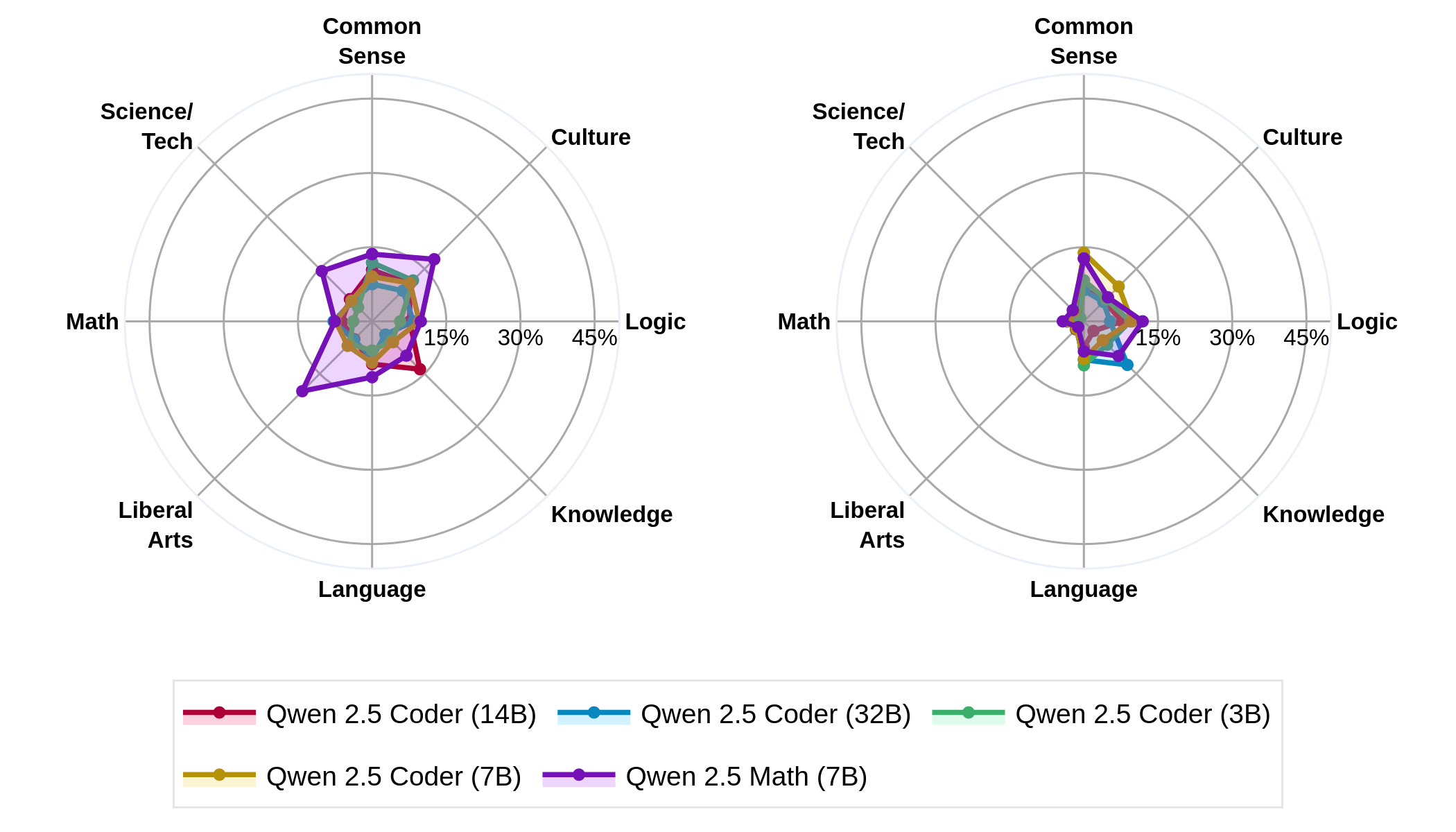}
  \caption{\textbf{Forgetting (left) and Backward Transfer (right) after domain‑continual pretraining.} Forgetting is low-to-moderate and consistent across categories; backward transfer is low. Scaling model size reduces forgetting.}
  \label{fig:dacp-training}
\end{figure}

\textbf{Motivation.} A popular class of continual learning works adapt general LLMs at the application layer for domains such as coding, mathematics, search, and tool use. As generalist LLMs are increasingly wrapped with tools and domain‑specific interfaces, specialization must not erode broad pretraining knowledge. Models still need to contextualize domain outputs, communicate with diverse users, respect cultural norms, and uphold safety and ethical standards. These needs motivate our study of forgetting and backward transfer under domain‑continual pretraining.

\textbf{Setup.} We study continual pretraining that converts a general base model into a specialized one, exemplified by Qwen2.5‑Coder~\citep{hui2024qwen2.5} and Qwen2.5‑Math~\citep{yang2024qwen25mathtechnicalreportmathematical}.\footnote{We treat domain‑continual  reasoning via SFT/RL separately in \S\ref{subsec:sft_rl} and focus on domain-continual training here.} Unlike general instruction tuning or reasoning post-training, domain-continual pretraining shifts the underlying representation using large, relatively uncurated, web‑scale domain corpora.

\textbf{Main results.} Figure~\ref{fig:dacp-training} summarizes our findings. Domain-continual pretraining induces little to moderate amounts of forgetting among all post‑training methods we evaluate. Backward transfer to general abilities is weak: Gains in the specialized domain rarely improve non‑target tasks. The effect spans categories of pretraining knowledge, with no single category driving it, although math‑specialized models show significantly more forgetting. Lastly, larger models forget less and have marginally better backward transfer.

\textbf{Qualitative analysis.} We performed manual errors analysis, which indicates reduced instruction‑following fidelity (e.g., weaker adherence to constraints, formats, and role‑specific directives). Evidence of this is found in supplemental tests, where a zero-shot, chat-template evaluation is done. In this case, a coder model may, for example, answer “Who was the president of the US?”  with a response followed by code, often with embedded answers, making extraction difficult of the "true answer". While few-shot prompting alleviates this, it demonstrates a weakened instruction-following ability and less easily elicited knowledge.

\takeaway{Domain‑continual pretraining yields low-to-moderate forgetting across categories; backward transfer is limited. Scaling model size marginally reduces forgetting. This indicates current domain-continual pretraining pipelines appears to alleviate much of the large forgetting behavior seen in previous literature.}

\vspace{-0.15cm}
\subsection{Subarea 2: Instruction Tuning}
\vspace{-0.15cm}
\label{subsec:it}

\begin{figure}[h]
  \centering
  \vspace{-0.3cm}
  \includegraphics[width=0.95\textwidth,keepaspectratio]{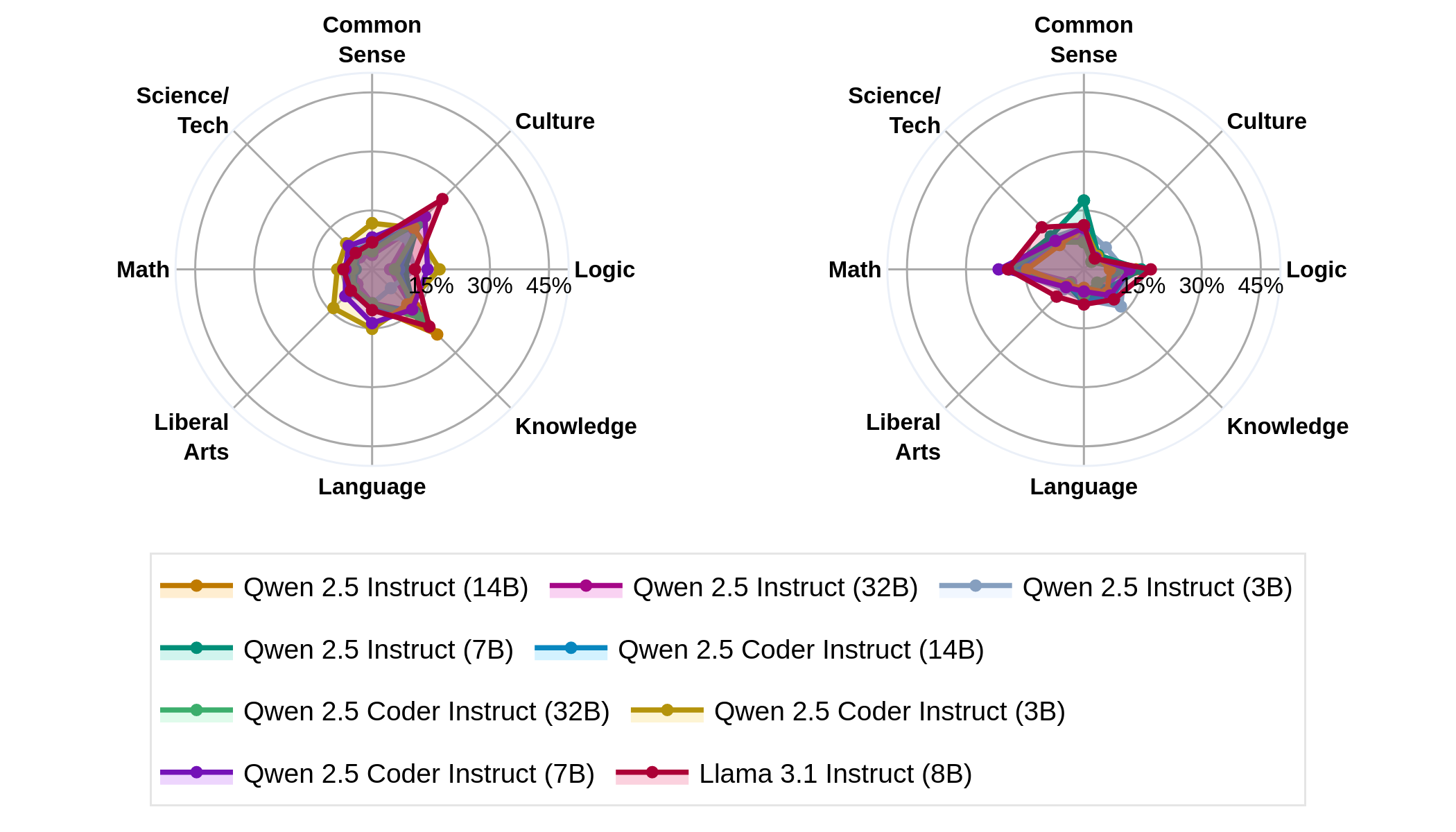}
  \caption{\textbf{Forgetting (left) and Backward Transfer (right) after instruction-tuning} yields moderate forgetting and backward transfer categories-wise. Scaling model size reduces forgetting and backward transfer.}\vspace{-0.3cm}
  \label{fig:instruction-tuning-training-base}
\end{figure}

\textbf{Motivation.} Base models often require carefully engineered prompts to elicit pretraining knowledge, limiting usability. Modern post‑training pipelines therefore add instruction tuning to enable natural user interaction with minimal prompting. Most continual‑learning work we surveyed focuses on mitigating forgetting in this setting. We ask: To what extent does instruction following come at the expense of previously learned knowledge?

\textbf{Setup.} We measure forgetting and backward transfer from instruction tuning in generalist models (Qwen2.5~\citep{qwen2.5}, Llama~3.1~\citep{llama3}) and domain‑continual pretrained models (Qwen2.5‑Coder)\footnote{Qwen2.5-Math Instruct is surprisingly tuned with GRPO which leads to it being classified under Reasoning}.

\textbf{Results.} As shown in Figure~\ref{fig:instruction-tuning-training-base}, there is low to moderate forgetting across models, with spikes in the Culture and Knowledge categories. However, there is substantial backward transfer in the Math category. Furthermore, scaling model size reduces forgetting and increases backward transfer. This effect is consistent across domain-general and domain-specific base models. While most of the continual learning literature focuses on reducing forgetting in this area, we note the forgetting is low to moderate with current training practices.

\textbf{Qualitative analysis.} Transfer gains likely reflect better elicitation of pretraining knowledge: Instruction‑tuned models use what they already know with straightforward prompts used in benchmarks, whereas base models often require carefully crafted prompts.

\takeaway{Instruction tuning produces low-to-moderate forgetting overall and moderate backward‑transfer, particularly in math, across model families; the forgetting and back-transfer tend to decrease with increasing model scale. Shifting focus to other subareas of post-training might spur interesting research directions, but there is still progress to be made in this area.}

\vspace{-0.15cm}
\subsection{Subarea 3: Training with Reasoning Traces (SFT and RL)}
\vspace{-0.15cm}
\label{subsec:sft_rl}

\textbf{Motivation.} Recent methods encourage explicit reasoning by letting models \textit{think} on a scratchpad before answering; which is now scaled in size and trace length with RL objectives. As training domains and data grow, we measure how much such reasoning training induces forgetting to guide continual‑learning practice.

\textbf{Setup.} We consider two settings: (i) starting from a base model and (ii) starting from an instruction‑tuned model. For the latter, we separate light‑touch post‑training (small datasets) from heavy post‑training. We do not separate RL from SFT as the behavior across forgetting and backward transfer is similar between the two objectives.

\subsubsection{Training with Reasoning Traces from Base Models}

\begin{figure}[htbp]
  \centering
  \vspace{-0.15cm}
  \includegraphics[width=0.95\textwidth,keepaspectratio]{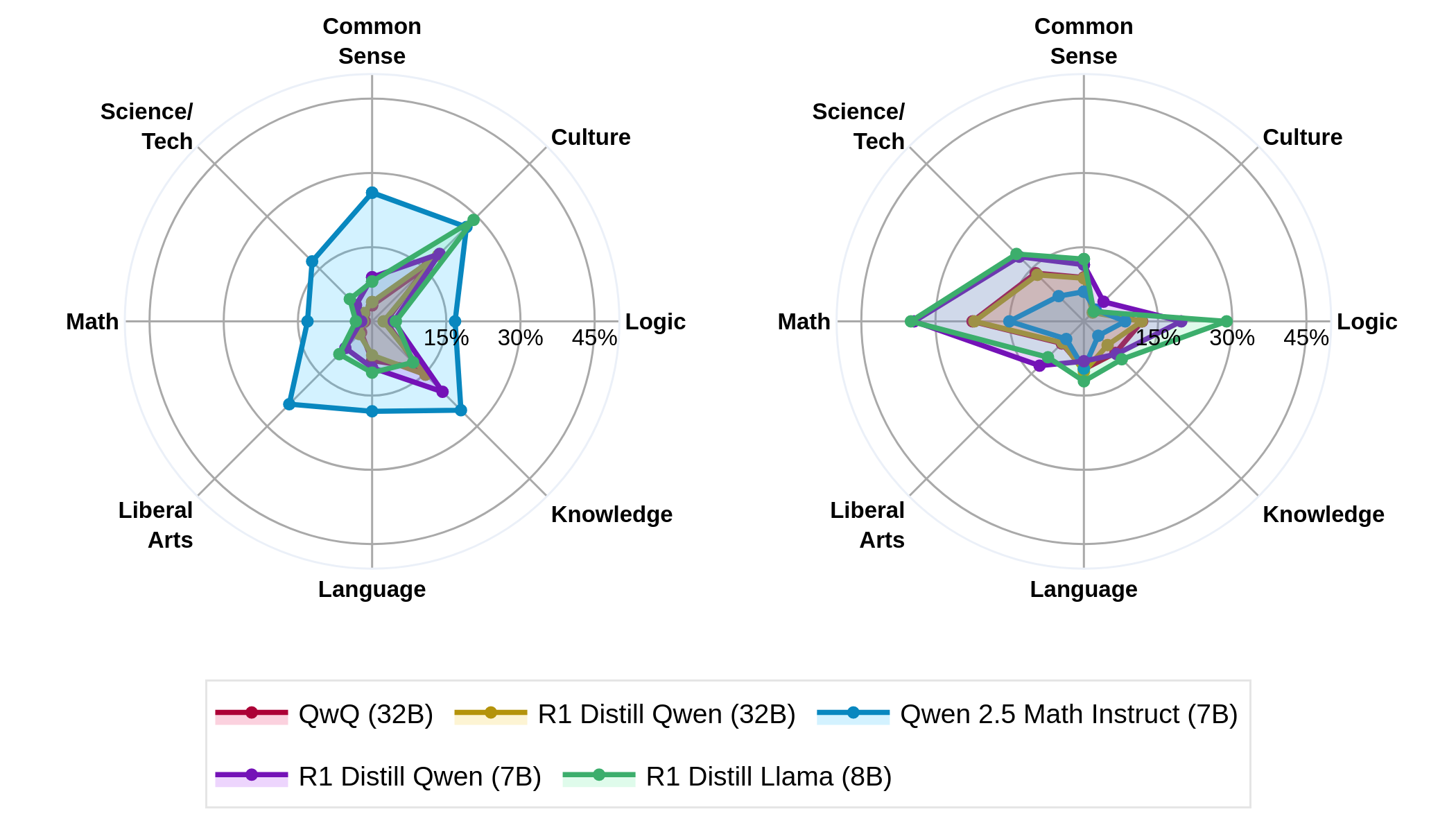}

  \caption{\textbf{Forgetting (left) and Backward Transfer (right) after reasoning training (SFT/RL) from base model.} It generally yields minimal forgetting, except in the Culture and Knowledge categories, and has moderate to high backward‑transfer gains. Qwen2.5 Math Instruct (7B) is an exception to this trend, demonstrating forgetting across all categories.}
\label{fig:reason-from-base-training}
\end{figure}

\textbf{Models.} We evaluate QwQ‑32B (from Qwen2.5‑32B Base)~\citep{qwq32b}, Qwen2.5‑Math‑7B‑Instruct (RL post‑trained with GRPO), and DeepSeek‑R1‑Distill models across different models (Qwen2.5 Base and Llama 8B base)~\citep{deepseekai2025}.

\textbf{Results.} From Figure~\ref{fig:reason-from-base-training}, we see that across scales, model families, and training types, we observe large gains, particularly in Math and Logic, in backward transfer with minimal forgetting. Forgetting is generally low, but is moderate for knowledge and large for Culture. The exception to this trend is the Qwen2.5 Math Instruct model which shows substantial forgetting across many categories. Sample-wise inspection shows this is primarily due to weak adhearance to the prompt, sometimes outputting random multi-lingual text. Except for this case, when compared to instruction tuning on the same base model (Figure~\ref{fig:instruction-tuning-training-base}), we see similar forgetting and larger back-transfer~\footnote{All corresponding tables are available in Appendix~\ref{sec:reasoning_from_base} for detailed comparison.}.

We conclude that much of the backward transfer reflects improved instruction following. To isolate reasoning effects beyond elicitation, the next sections analyze reasoning training that starts from an instruction‑tuned model, for better exploration of gains. However, models with light-touch reasoning training (i.e. low data) behave differently from those trained at scale (i.e. high data).  We therefore present these two cases separately.

\takeaway{Training with SFT/RL for reasoning results in dynamics similar to instruction tuning, but to an even greater extent: We generally observe low to moderate forgetting overall and larger category-specific backward transfer gains. Forgetting mitigation in this domain should consider broad categories of knowledge/abilities when measuring forgetting and back transfer.}

\subsubsection{Reasoning Training from Instruction-Tuned Models: Low-Data Scenario}

\begin{figure}[htbp]
  \centering
  \vspace{-0.1cm}
  \includegraphics[width=0.95\textwidth,keepaspectratio]{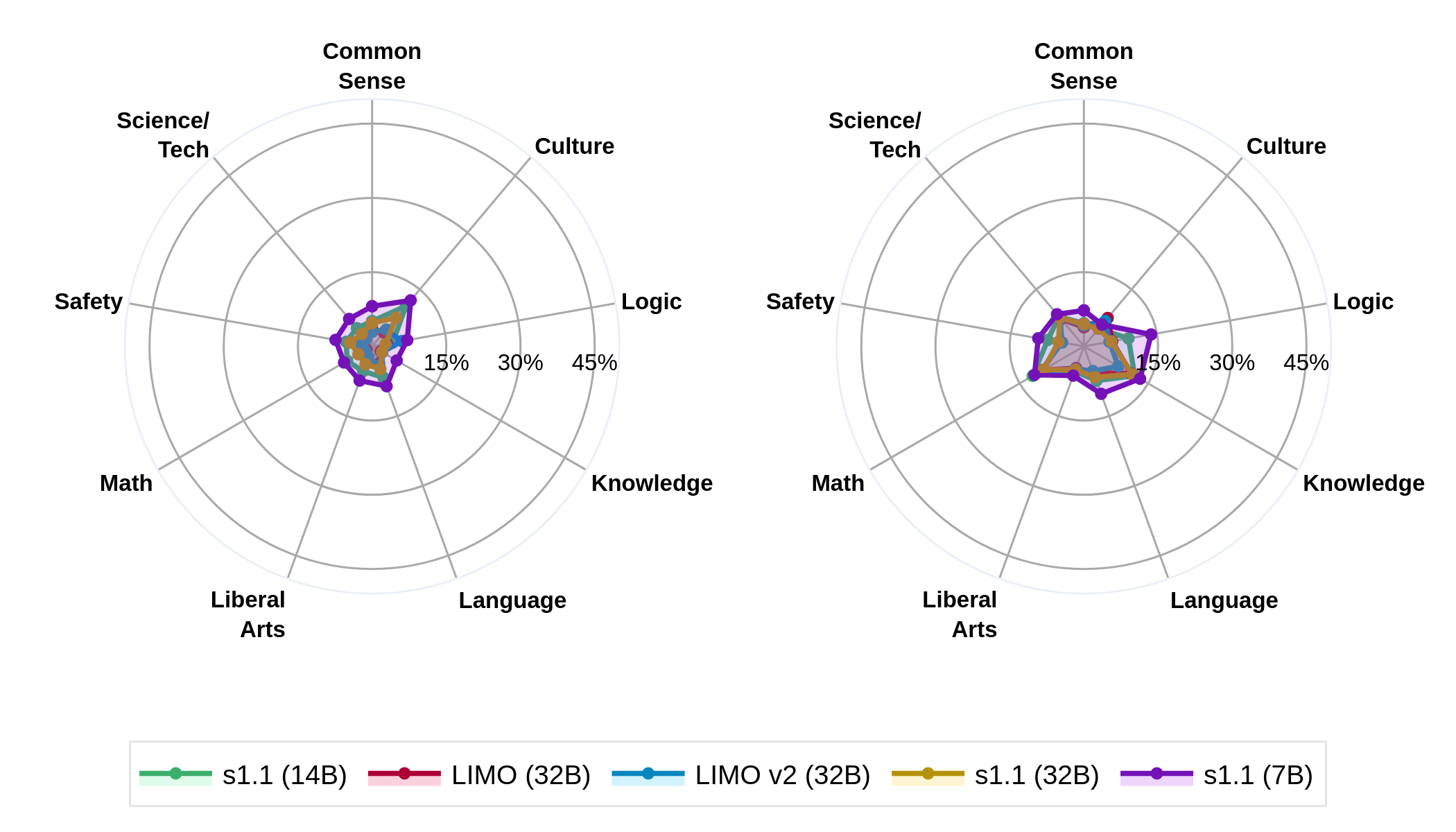}
  \caption{\textbf{Forgetting (left) and Backward Transfer (right) after reasoning training from instruct: low data scenario.} Yields little forgetting and backward transfer. Forgetting decreases with model scale.}\vspace{-0.3cm}
\label{fig:low-data-from-instruct-training}
\end{figure}

\textbf{Models.} We use the s1.1 family (7B, 14B, 32B)~\citep{muennighoff2025s1simpletesttimescaling} and LIMO (v1 and v2)~\citep{ye2025limoreasoning} all tuned from corresponding sized Qwen instruct models.

\textbf{Results.} Figure~\ref{fig:low-data-from-instruct-training} summarizes our findings. Across categories, models show minimal forgetting and low backward transfer. This makes sense, as training for a few passes on little data leaves pretraining knowledge largely intact. That is, the model does not forget much, but it also exhibits little backward transfer gains beyond the instruction‑tuned baseline. Scaling model size marginally lowers forgetting, and the smaller teacher–student gap similarly tends to reduce backward transfer, with the exception of the Knowledge category.

\takeaway{For low‑data regime, reasoning training from instruct models yields low forgetting and backward transfer. Forgetting decreases with model scale; backward transfer gains also tend to fall with a narrowing student-teacher gap. This suggests that future forgetting mitigation literature on reasoning models should focus on medium-to-large sized training datasets.}

\vspace{-0.15cm}
\subsubsection{Reasoning Training from Instruction-Tuned Models: High-Data Scenario}
\vspace{-0.15cm}

\begin{figure}[H]
  \centering
  \vspace{-0.3cm}
  \includegraphics[width=0.95\textwidth,keepaspectratio]{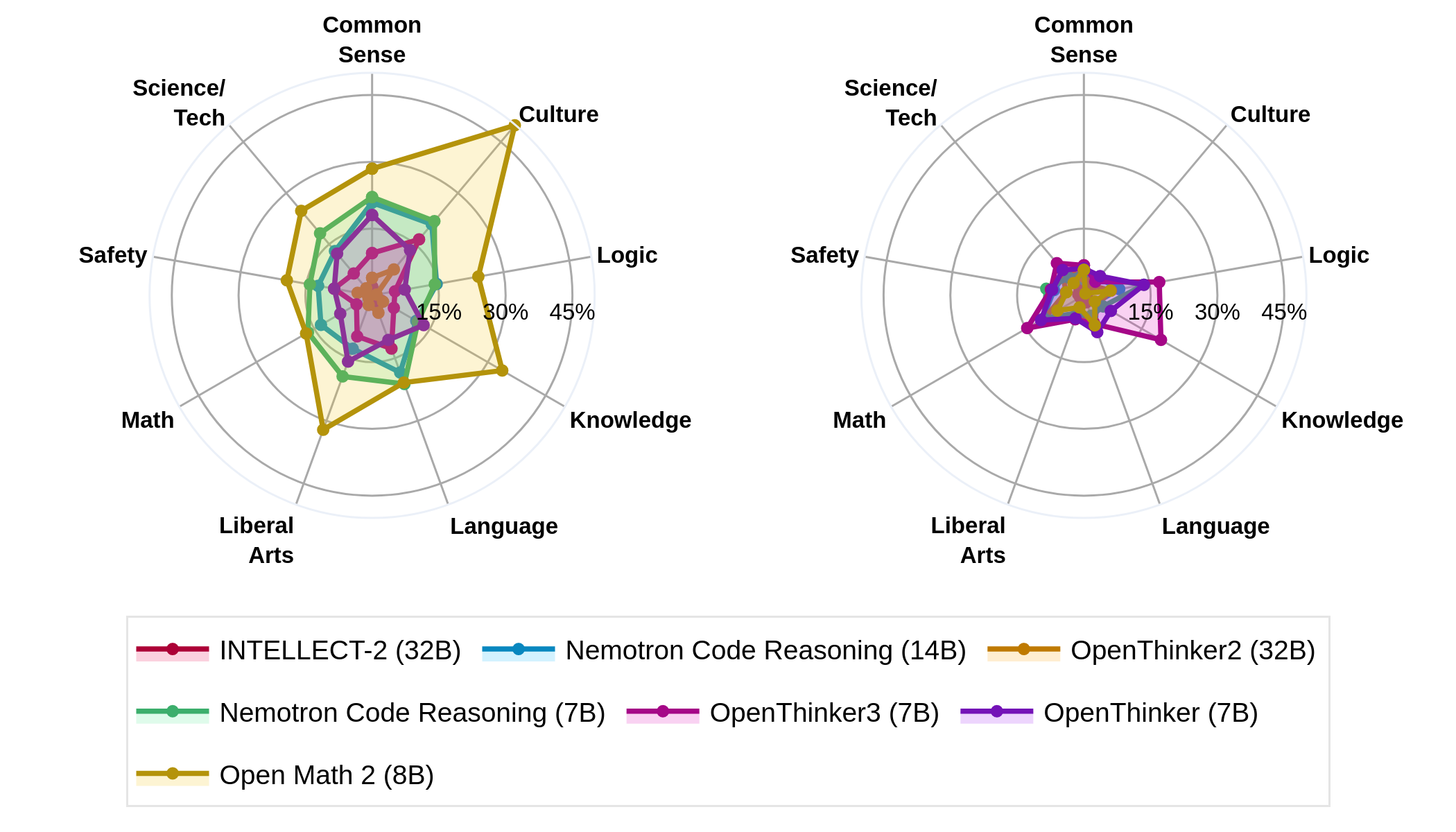}
  \caption{\textbf{Forgetting (left) and Backward Transfer (right) after reasoning training from instruct: high data scenario.} No single factor robustly explains the dynamics of forgetting and backward transfer.}\vspace{-0.3cm}
\label{fig:high-data-from-instruct-training}
\end{figure}

\textbf{Models.} We evaluate OpenCodeReasoner and OpenMath2~\citep{bercovich2025llamanemotronefficientreasoningmodels}, OpenThinker-7B, Openthinker2-32B, and OpenThinker3‑7B~\citep{guha2025openthoughtsdatarecipesreasoning}, and Intellect‑2‑32B~\citep{primeintellectteam2025intellect2reasoningmodeltrained}. This spans SFT (former) and RL (Intellect‑2).

\textbf{Results.} Results vary by domain mix and model quality.  The OpenThinker models shows low–to–moderate forgetting and moderate backward transfer, perhaps due to the breadth of the training datamix, whereas OpenCodeReasoner models show consistently high forgetting with low backward transfer gains due to the narrower training data. Furthermore, we find this may be primarily  due to weakened instruction-following capabilities, as sample-level inspection shows the model will refuse to answer with letters, when numbers are present as options, instead answering numerically. This is also seen with the Nemotron Code Reasoning models, where answers will often be embedded within python code. These factors make the forgetting and backtransfer observed highly dependent on the extraction method used. Scaling model size, if compared in OpenThinker models, signals improvements in both forgetting and backward transfer -- as seen in most previous sections. Decentralized training (as in Intellect-2), in contrast, showed minimal forgetting or backward transfer. We conjecture that the model largely remain unchanged compared to the original model as it shows negligible gains on the optimized math benchmarks~\cite{hochlehnert2025sober}. However, the results here remain preliminary. We do not find a single dominant factor—initialization, data regime, or scale that sufficiently explains forgetting and backward‑transfer dynamics. We believe controlling the finer details which determine the quality of the trained model might lead to better conclusions.

\takeaway{No single factor robustly explains the dynamics of forgetting and backward transfer -- training on a mix of domains appear to improve both forgetting and backward transfer.}

\section{Does Model Merging Reduce Forgetting?}
\begin{figure}[htbp]
  \centering
  \vspace{-0.3cm}
  \includegraphics[width=0.95\textwidth,keepaspectratio]{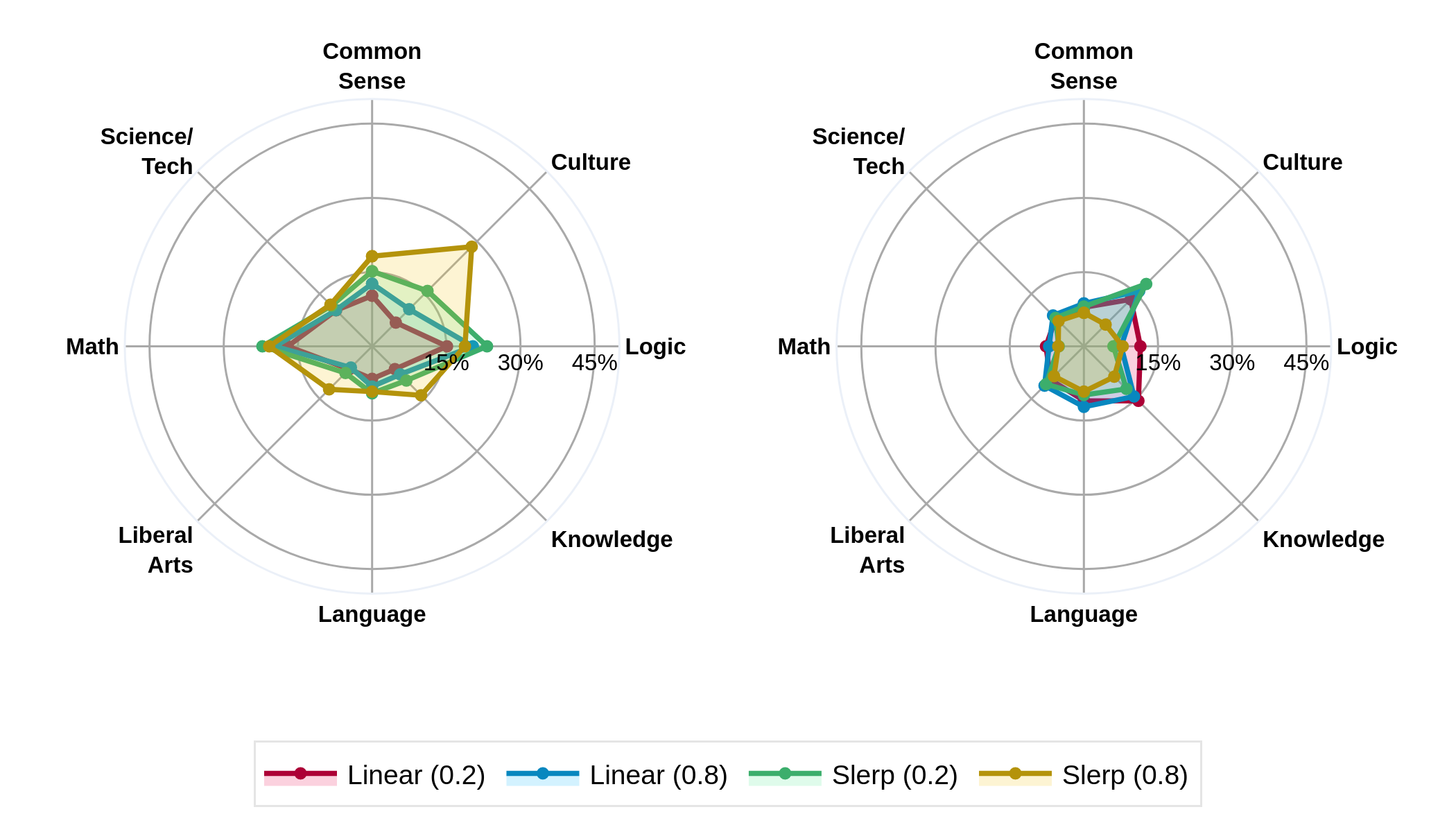}
  \caption{\textbf{Forgetting and Backward Transfer of Qwen 2.5 Base merged with Qwen 2.5 Coder (7B) relative to Qwen2.5 Coder}. Induces moderate forgetting and little backward transfer.}\vspace{-0.3cm}
\label{fig:merge-qwen-coder-after}
\end{figure}

\textbf{Motivation.} Recent work shows that offline model merging can combine capabilities from multiple models~\citep{dziadzio2025merge}. Unlike classical continual learning~\citep{delange2022}, it requires neither the original training data nor the ability to resume training, which is practical in resource‑constrained settings.

\textbf{Setup.} We evaluate Exponential Moving Average (EMA) merging; in the two‑checkpoint case this is linear interpolation,
\[
\theta_{\text{EMA}}(\alpha) = \alpha\,\theta_{\text{pre}} + (1-\alpha)\,\theta_{\text{post}}.
\]
Prior large‑scale studies find these simple schemes effective for continual learning with foundation models~\citep{roth2024practitioner}. Our experiments compare linear interpolations (e.g. LERP and SLERP) across OpenThinker‑7B, OpenThinker3‑7B,  and Qwen2.5‑Coder‑7B, together with their base checkpoints.

\textbf{Results.} We compare merged checkpoints to the post‑trained model \(\theta_{\text{post}}\); results for \(\theta_{\text{pre}}\) appear in the Appendix. For Qwen2.5‑Coder‑7B and OpenThinker3‑7B, even small mixes with the base checkpoint degrade performance, severely for the latter case (Figures~\ref{fig:merge-qwen-coder-after}, \ref{fig:merge-openthinker3-after}). In contrast, OpenThinker‑7B shows small overall gains, accompanied by moderate forgetting (Figure~\ref{fig:merge-openthinker-after}). In our setting, merging does not mitigate forgetting. This may reflect that we merge only two checkpoints, whereas prior work often merges eight or more~\citep{yadav2023tiesmerging, yadav2024matters}. We further hypothesize that weight drift between our checkpoints is larger than is typical in the merging literature, which could explain these outcomes.

\takeaway{Model merging does not yet reliably mitigate forgetting in post‑training pipelines.}

Merging remains promising, but further study is needed to determine when each method works, how to overcome its limitations, and whether an increased scale can compensate for these difficulties. Future works may consider the effect of the number of models merged and weight drift on reasoning models.

\vspace{-0.15cm}
\section{Conclusion}
\label{sec:conclusion}
We present a new metric for sample-wise forgetting and backward transfer that corrects for chance in multiple-choice evaluations. Our results challenge a common claim: sequential training does not automatically erode pre-training knowledge. Forgetting depends on the post-training method and its scale. By focusing on sample-wise forgetting, we offer a clearer map of what knowledge is lost and in what stages of instruction tuning do language models lose during post-training -- providing fertile ground to study how to preserve (minimize forgetting) and accumulate (higher backward transfer) knowledge while adding new capabilities by post-training. Promising ways to prevent forgetting include: (1) Designing objectives and data that explicitly penalize 1→0 transitions; (2) Using targeted synthetic corpora or brief mid-training bursts to repair localized forgetting; (3) Adding retrieval mechanisms to reduce reliance on in-weight knowledge storage.

\subsubsection*{Acknowledgments}
AP and MB acknowledge financial support by Federal Ministry of Research, Technology and Space (BMFTR) FKZ: 16IS24085B and Open Philanthropy Foundation funded by the Good Ventures Foundation. AH acknowledges funding by the Federal Ministry of Research, Technology and Space (BMFTR), FKZ: 16IS24079A. AH thanks the International Max Planck Research School for Intelligent Systems (IMPRS-IS) for support.

\bibliography{iclr/main}
\bibliographystyle{iclr/iclr2026_conference}

\newpage
\appendix
\part{Appendix}

\localtableofcontents
\clearpage
\section{Extended Related Works}

\paragraph{Post-training techniques.} A broad set of post-training methods now underpins standard LLM pipelines. \emph{Supervised fine-tuning (SFT)}~\citep{ouyang2022training} remains the core step, used for continued pre-training and instruction tuning. At later stages, \emph{reinforcement learning from human feedback (RLHF)}~\citep{ouyang2022training} aligns model outputs with human preferences. To simplify preference learning, \emph{direct preference optimization (DPO)}~\citep{rafailov2023direct} provides a direct loss surrogate. With the rise of test-time scaling (e.g. sampling depth or compute at inference), \emph{group relative policy optimization (GRPO)}~\citep{shao2024deepseekmathpushinglimitsmathematical} has been proposed to elicit stronger intrinsic reasoning. Taken together, these methods introduce distinct objectives and optimizers, increasing the complexity of the post-training stack~\citep{wang2025uftunifyingfinetuningsft}.

\paragraph{Measuring catastrophic forgetting.} Catastrophic forgetting is the loss of previously acquired knowledge when a network learns new information. Early studies examined the effect in small models and simplified settings~\citep{mccloskey1989,ratcliff1990,french1999}. \citet{lopez2017gradient} formalized forgetting via \emph{backward transfer}, the effect of learning a new task on performance in earlier ones: positive values indicate improvement; negative values indicate forgetting. Recent work extends these analyses to deep networks trained on large-scale data, with growing attention to language models~\citep{biesialska-etal-2020-continual,wu2022pretrained}.

\paragraph{Benchmark paradigm.} Task-incremental learning is the dominant paradigm for benchmarking forgetting~\citep{delange2022}. Models learn a sequence of tasks with clear boundaries, and task labels are available at train and test time. Class-incremental learning removes test-time task identifiers, making evaluation stricter~\citep{wang2024}. Other views analyze continual learning through positive/negative transfer~\citep{yildiz2025}. At the sample level, \citet{toneva2018an} introduced forgetting metrics that identify “unforgettable” examples (stable once learned) and “catastrophically forgotten” examples (highly plastic), and showed these patterns are consistent across architectures and random seeds.

\paragraph{Language-model forgetting.}
Recent studies focus on forgetting induced by instruction tuning. \citet{luo2025empiricalstudycatastrophicforgetting} trained models up to $7$B parameters with SFT and evaluated multiple knowledge categories. \citet{deepseek-llm} reported instruction-tuning-related regressions on sentence completion even for $67$B models. \citet{fernando2025mitigatingforgettingllmsupervised} examined forgetting across SFT followed by RLHF and proposed joint-training strategies to mitigate it. \citet{lin2024mitigatingalignmenttaxrlhf} framed instruction-tuning degradation as an “alignment tax” (performance loss on pre-training skills due to alignment) and found model merging to be the most Pareto-efficient mitigation among tested techniques. \citet{li2024examiningforgettingcontinualpretraining} studied continual pre-training on aligned LMs and observed notable regressions in alignment-related behavior.

\paragraph{Catastrophic forgetting in reasoning training pipelines.} Work on reasoning-oriented LMs highlights new failure modes. \citet{li2025temporal} defined \emph{temporal forgetting}: models lose the ability to solve problems they could solve at earlier training checkpoints. The effect appears in both RL-trained and instruction-tuned models. They proposed \emph{temporal sampling}—round-robin sampling from recent checkpoints—as a mitigation. \citet{pipatanakul2025adaptinglanguagespecificllmsreasoning} merged a language-fine-tuned model with DeepSeek R1 Distill ($70$B; both derived from Llama~3.3 $70$B~\citep{llama3}) to adapt reasoning while preserving language competence. For multimodal models, \citet{chen2025bring} found that later layers primarily support reasoning, whereas early layers concentrate perception, suggesting layer-wise interventions. We document forgetting extensively across post-training pipelines in our work.

Each new method introduces its own objective and optimization procedure, adding to the complexity of the post‑training landscape~\citep{wang2025uftunifyingfinetuningsft}.

\paragraph{Mitigation strategies.} Sequential SFT to RLHF/DPO can exacerbate forgetting. To counteract this, researchers explore: (i) \emph{model averaging}, interpolating between pre- and post-RLHF checkpoints to trade off alignment and retention~\citep{lin2024mitigatingalignmenttaxrlhf}; (ii) \emph{joint post-training}, optimizing supervised and preference objectives simultaneously with convergence guarantees~\citep{fernando2024mitigating}; and (iii) \emph{unified fine-tuning (UFT)}, which folds instruction tuning and alignment into a single implicit-reward objective~\citep{wang2025uftunifyingfinetuningsft}. Additional techniques—including advantage models and selective rehearsal—stabilize RLHF by shaping reward distributions and replaying curated data~\citep{peng2023stabilizingrlhfadvantagemodel}. \emph{Online Merging Optimizers (OMO)} combine gradients from SFT and RLHF models during training to maximize reward while preserving pre-trained skills~\citep{lu2024online}. Theory supports these interventions: up to permutation symmetries, weights of homologous models tend to lie in a shared low-loss basin~\citep{ainsworth2023git}. Hence, we were quite surprised that model merging does not work for our simple case of mitigating forgetting during post-training with only two deep networks.

\paragraph{Forgetting at scale.} Pre-training mitigates forgetting relative to training from scratch~\citep{mehta2023,McRae1993}. \citet{ramasesh2022effect} further found that pretrained ResNets and Transformers (up to $\sim$$100$M parameters) are robust to forgetting at scale; language experiments showed similar trends. However, \citet{luo2025empiricalstudycatastrophicforgetting} reported increased forgetting with scale in the $1$–$7$B LM regime, suggesting modality- and regime-dependent behavior. In contrast to these works, we study forgetting during post-training of language models.

\section{Experimental Setup}
\subsection{Evaluation}
To evaluate performance differences between models, we employ chain-of-thought (CoT) prompting \cite{wei2022chain} on multiple-choice question answering (MCQA) datasets. In this setup, the model auto-regressively generates a reasoning chain prior to producing its final answer. The predicted choice is then extracted from the generated text and compared against the ground-truth label. When available, chat-specific templates are incorporated into the prompt to ensure consistent formatting.

Because some models, particularly base models, tend to continue generating responses for subsequent questions after completing the current one, we provide explicit stop sequences to terminate generation once a prediction has been produced.

To encourage answers in strict MCQA format (models sometimes output the option text instead of the letter), we prepend the following instruction prompt:

\begin{verbatim}
{Instruction}

On the very last line, write exactly "Answer: $LETTER" (e.g. 
"Answer: B"), with no extra punctuation, no lowercase, no *, 
and no trailing spaces.
Think step by step, showing your reasoning.
Question: "{Question}"
\end{verbatim}

For the case of base models, where few-shot prompting yields a more accurate elicitation of their knowledge, we use few-shot prompting:
\begin{verbatim}
{Instruction}

Question: "{Few-Shot Question 1}"
Reasoning: {Few-shot Reasoning Trace 1}
Answer: {Few-shot Answer 1}

... <--- more examples

Question: "{Question}"
Reasoning: 
\end{verbatim}

Datasets where CoT reasoning traces are provided for few-shot prompting, we use those. In the cases where this is not provided (PIQA, MCTest, Social-IQa, ARC, MCTest, and Hellaswag) CoT few-shot examples were generated and then confirmed these are not included in the benchmarks\footnotemark[1].

All experiments are conducted using the Hugging Face \texttt{LightEval} framework, with results logged at the sample level. For generation, we allow up to 32{,}768 tokens, which we found sufficient for models to complete their chain of thought and provide an answer. In cases where the maximum trained context length is smaller, then the generation is reduced to that number, as is the case with Qwen2.5-$7$B-Math and Qwen2.5-$7$B-Math-Instruct \cite{qwen2.5}. The temperature is set to 0.6 and nucleus sampling with $p=0.95$ is applied.

\subsection{Datasets}

To evaluate broad model knowledge and capabilities, we benchmark on twelve public datasets: MMLU \cite{hendryckstest2021, hendrycks2021ethics}, BBH \cite{suzgun2022challenging}, GPQA \cite{rein2024gpqa}, MuSR \cite{sprague2024musr}, ARC \cite{allenai:arc}, TruthfulQA \cite{lin-etal-2022-truthfulqa}, HellaSwag \cite{zellers2019hellaswag}, Social IQa \cite{sap-etal-2019-social}, MCTest \cite{richardson-etal-2013-mctest}, PIQA \cite{Bisk2020}, CommonsenseQA \cite{talmor-etal-2019-commonsenseqa}, and SaladBench \cite{li2024salad}. Several of these benchmarks, namely MMLU and BBH provide subject-level annotations, enabling fine-grained sub-benchmark analyses in addition to aggregate reporting. For the cases of MMLU and BBH, subcategory labels are provided which allow for splitting into further sub-benchmark evaluates by subjects. To enable easier understanding, we group these (sub-)benchmarks into high-level groups used to evaluate the capabilities of the models. They are grouped such that (sub-)benchmarks in the same group show similar trends in forgetting and improvement.

They are grouped as follows:

\textbf{Commonsense:}
\begin{itemize}
	\item Commonsense QA
	\item PIQA
\end{itemize}

\textbf{Culture:}
\begin{itemize}
	\item BBH (sports understanding and movie recommendation)
\end{itemize}

\textbf{Logic}
\begin{itemize}
	\item BBH (navigate, causal judgment, penguins in a table, web of lies, tracking shuffled objects three objects, tracking shuffled objects seven objects, tracking shuffled objects five objects, temporal sequences, reasoning about colored objects, logical deduction three objects, logical deduction seven objects, logical deduction five objects, formal fallacies, and date understanding)
	\item ARC (easy and challenge)
	\item MuSR (murder mysteries, object placements, and team allocation)
	\item MMLU (logical fallacies)
\end{itemize}

\textbf{Knowledge}
\begin{itemize}
	\item BBH (object counting)
	\item MMLU (miscellaneous and global facts)
	\item MCTest
\end{itemize}

\textbf{Language}
\begin{itemize}
	\item BBH (snarks, disambiguation qa, ruin names, and hyperbaton)
	\item Social IQa
	\item Hellaswag
    \item BBH (salient translation error detection)
\end{itemize}

\textbf{Liberal Arts}
\begin{itemize}
	\item MMLU (world religions, us foreign policy, sociology, security studies, public relations, professional psychology, professional law, prehistory, philosophy, management, international law, high school world history, high school us history, high school psychology, high school microeconomics, high school macroeconomics, high school government and politics, high school geography, and high school european history)
\end{itemize}

\textbf{Math}
\begin{itemize}
	\item BBH (geometric shapes, and boolean expressions)
	\item MMLU (high school statistics, high school mathematics, formal logic, elementary mathematics, econometrics, college mathematics, and abstract algebra)
\end{itemize}

\textbf{Safety \footnotemark[2]}
\begin{itemize}
	\item MMLU (moral scenarios, moral disputes, jurisprudence, and business ethics)
	\item TruthfulQA (mc1)
	\item SaladBench (mrq)
\end{itemize}

\textbf{Science \& Tech}
\begin{itemize}
	\item MMLU (marketing, virology, professional medicine, professional accounting, nutrition, medical genetics, machine learning, human sexuality, human aging, high school physics, high school computer science, high school chemistry, high school biology, electrical engineering, conceptual physics, computer security, college physics, college medicine, college computer science, college chemistry, college biology, clinical knowledge, astronomy, and anatomy)
	\item GPQA (diamond)
\end{itemize}

Unless otherwise noted, we follow the standard task formats and official evaluation splits; for TruthfulQA we report MC1, for GPQA the \emph{Diamond} subset, and for SaladBench the MRQ configuration. This taxonomy serves as the backbone for our analyses of capability acquisition and retention across training and deployment.

\footnotetext[1]{MMLU is evaluated with few-shot no CoT prompting for the base models}
\footnotetext[2]{These are only used in comparisons which do not include a base model because TruthfulQA and SaladBench are designed measure the default behavior of the model rather than knowledge, which few-shot prompting would bias.}

\clearpage
\section{Evaluation Comparison}
\subsection{Prompting Method}
In additional tests, we measure the ability of base-models using the same prompting as instruction-tuned models. Under these conditions we see ostensibly large forgetting in domain-continual pretrained models. Qualitative analysis suggests that this is largely due to the models outputting code, wherein the location of the answer can be obscured. When this is contrasted with the few-shot prompting, where there is much less forgetting, we conclude that forgetting metrics can vary significantly depending on the way knowledge is elicited, especially when training on narrow tasks.
\begin{figure}[h]
  \centering
  \includegraphics[width=\textwidth,keepaspectratio]{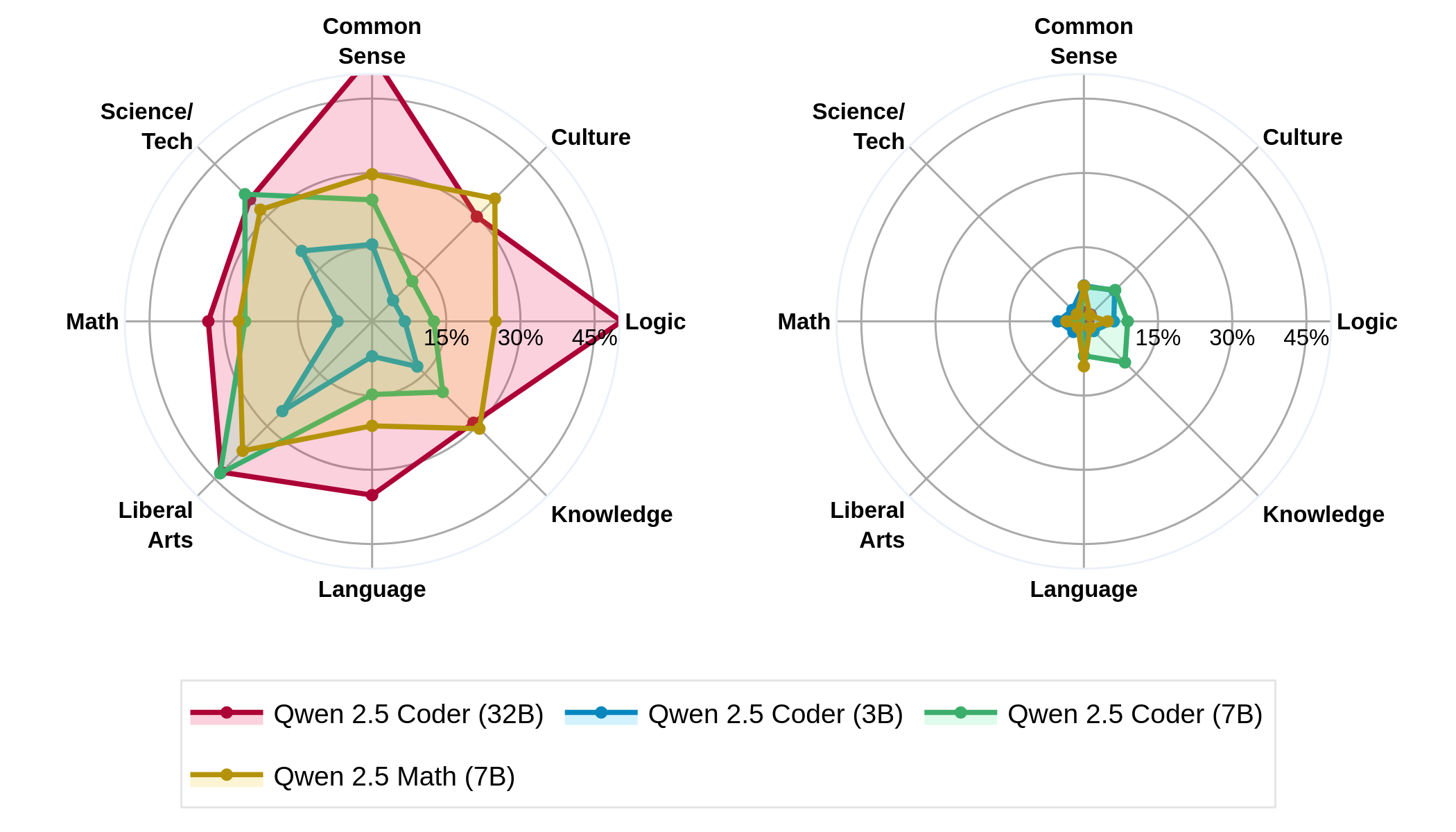}
  \caption{Domain-Adaptive Pretraining models using chat template prompting}
\end{figure}

Due to this, measuring performance of base models on certain datasets can become nontrivial. While benchmarks measuring knowledge or capabilities may be elicted through few-shot prompting, others, such as truthfulness or safety become more difficult as prompting them with examples would bias their behavior. Further works should consider exploring the effect of providing no-knowledge few-shot prompting, where the format of the question and answer is provided without leaking examples to avoid biasing the base model's output.
\clearpage

\subsection{Metric}

\begin{figure}[h]
  \centering
  \includegraphics[width=\textwidth,keepaspectratio]{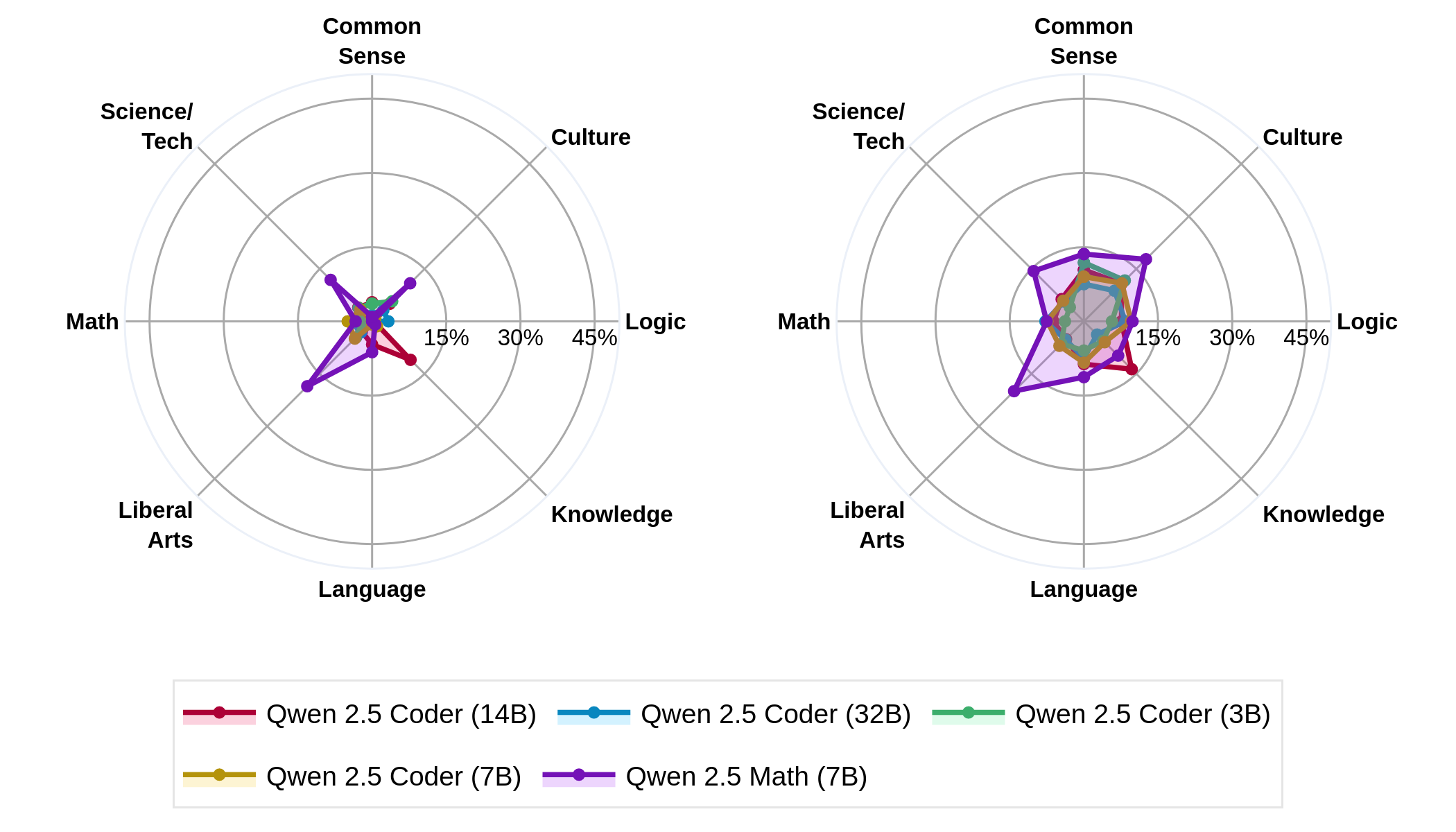}
  \caption{\textbf{Coder model comparing conventional forgetting (left) against our sample-wise forgetting (right).} More forgetting is uncovered when using the sample-wise forgetting metric.}
  \label{fig:metric_comparison}
\end{figure}
The sample-wise nature of the introduced metric allows more forgetting to be uncovered than is possible with the standard metric, the difference between the accuracy before training from the accuracy after training, clipped at 0. We demonstrate this on the example of the forgetting when undergoing domain-continual pretraining from Qwen2.5 to Qwen2.5 Coder (7B) when using the standard metric.

From Figure~\ref{fig:metric_comparison}, there appears to be little forgetting when using conventional forgetting. However, when using sample-wise data, moderate forgetting can be seen.

\clearpage

\section{Model Merging: Results}
\subsection*{Failure case: OpenThinker3}
\begin{figure}[htbp]
  \centering
  \includegraphics[width=\textwidth,keepaspectratio]{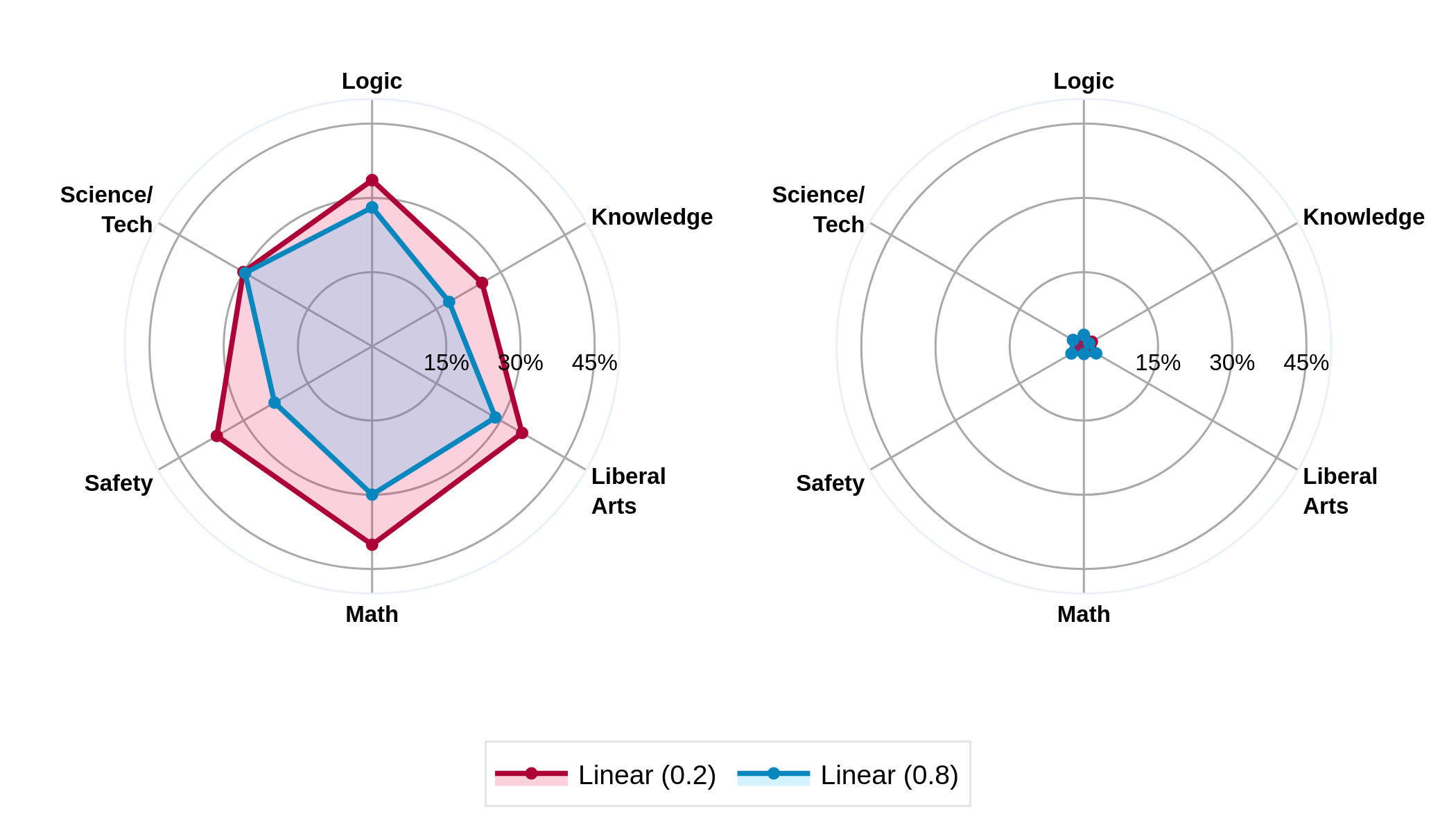}
  \caption{\textbf{Forgetting (left) and Backward Transfer (right) of Qwen 2.5 Instruct merged with OpenThinker3 (7B) relative to Qwen 2.5 Instruct on MMLU}. Large forgetting occurs. Sample-level analysis shows the model output degeneration, with the model often repeating words or phrases, typically without providing a final answer.}
\label{fig:merge-openthinker3-before}
\end{figure}

\begin{figure}[htbp]
  \centering
  \includegraphics[width=\textwidth,keepaspectratio]{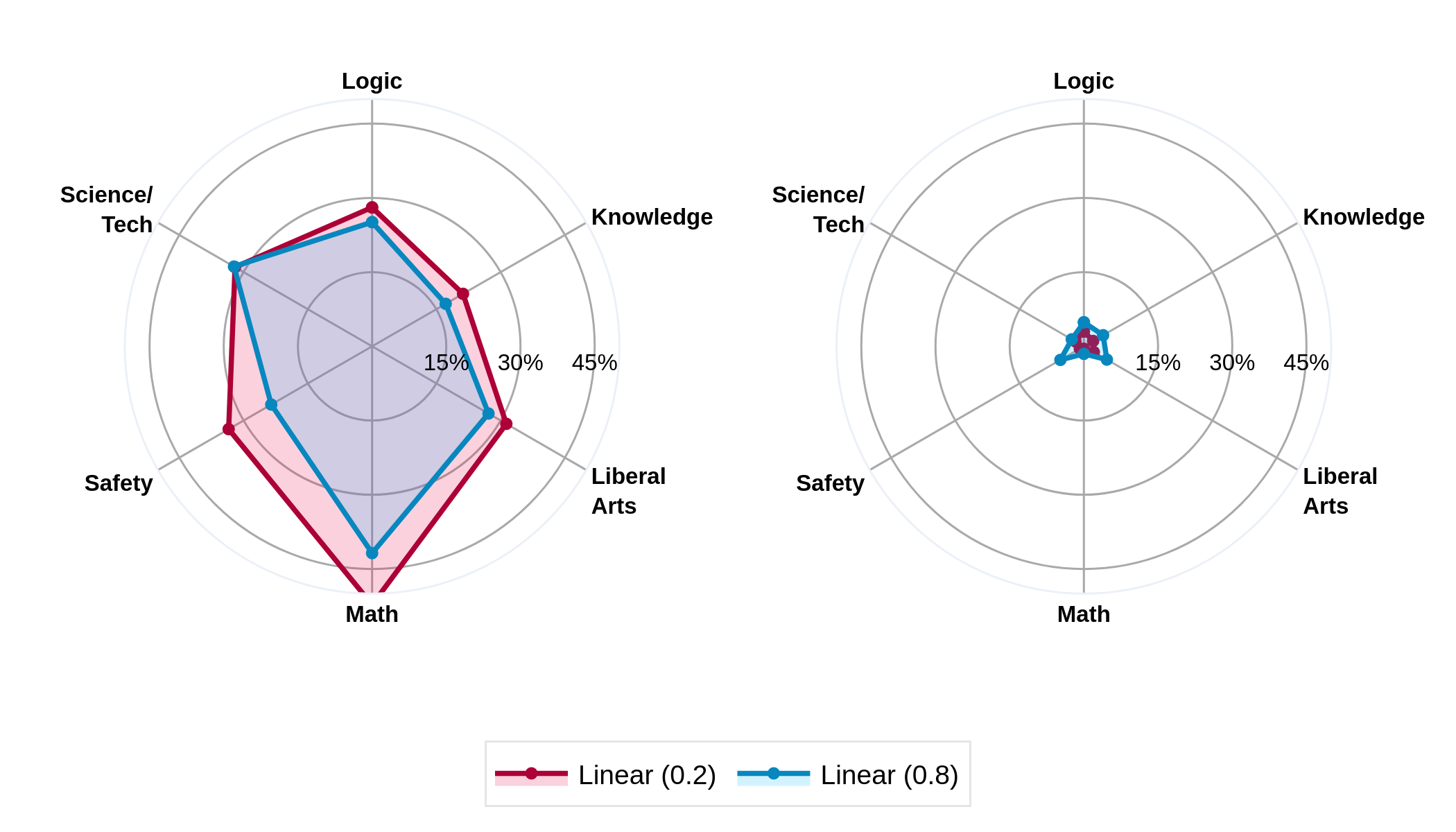}
  \caption{\textbf{Forgetting (left) and Backward Transfer (right) of Qwen 2.5 Instruct merged with OpenThinker3 (7B) relative to OpenThinker3 on MMLU}. Large forgetting occurs. Sample-level analysis shows the model output degeneration, with the model often repeating words or phrases, typically without providing a final answer.}
\label{fig:merge-openthinker3-after}
\end{figure}
\clearpage

\subsection*{Failure case: Coder Models}
\begin{figure}[htbp]
  \centering
  \includegraphics[width=\textwidth,keepaspectratio]{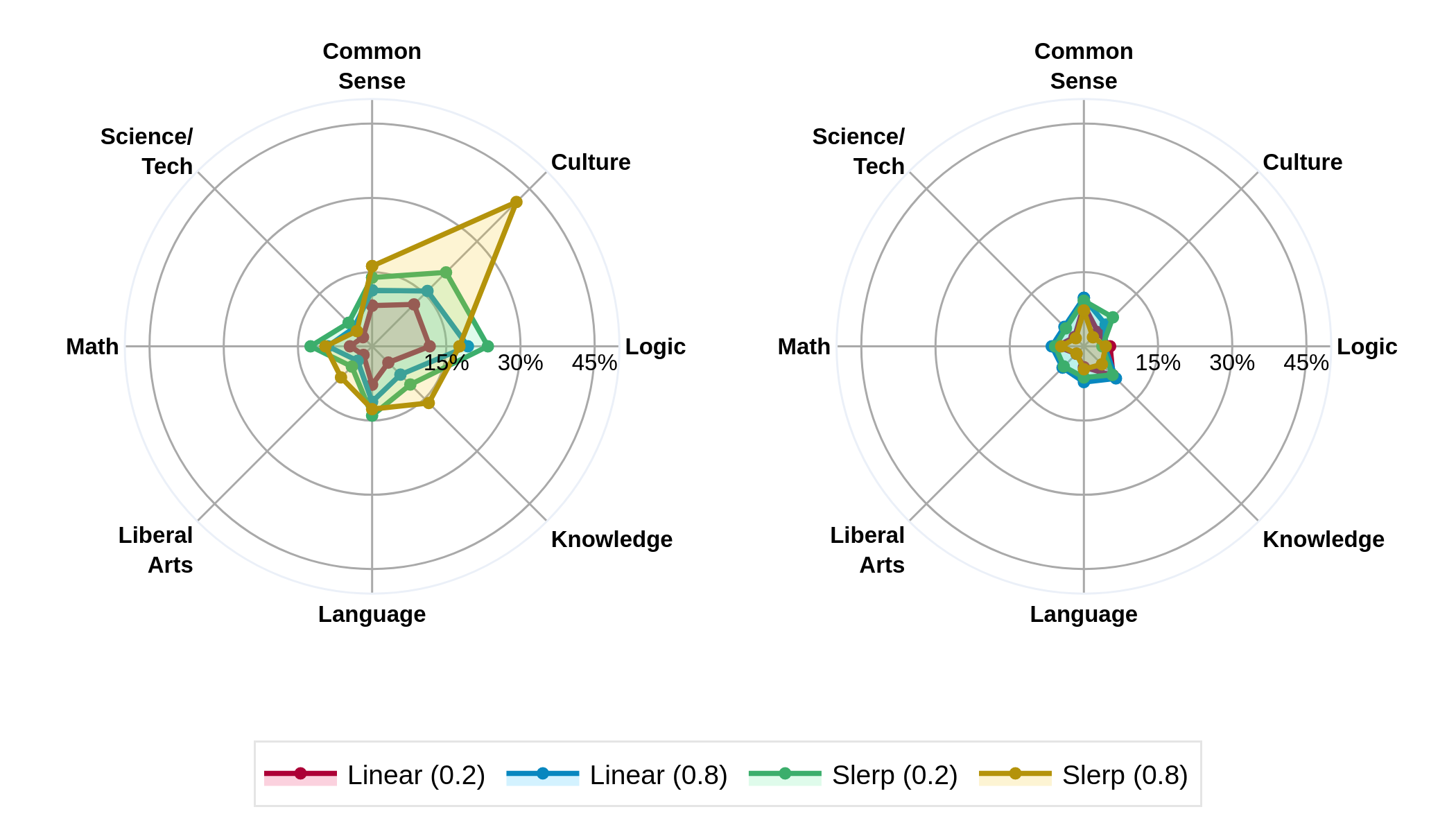}
  \caption{\textbf{Forgetting (left) and Backward Transfer (right) of Qwen 2.5 Base merged with Qwen 2.5 Coder (7B) relative to Qwen 2.5 Base on all Benchmarks}. Moderate-to-large forgetting occurs with low backward transfer.}
\label{fig:merge-coder-before}
\end{figure}

\begin{figure}[htbp]
  \centering
  \includegraphics[width=\textwidth,keepaspectratio]{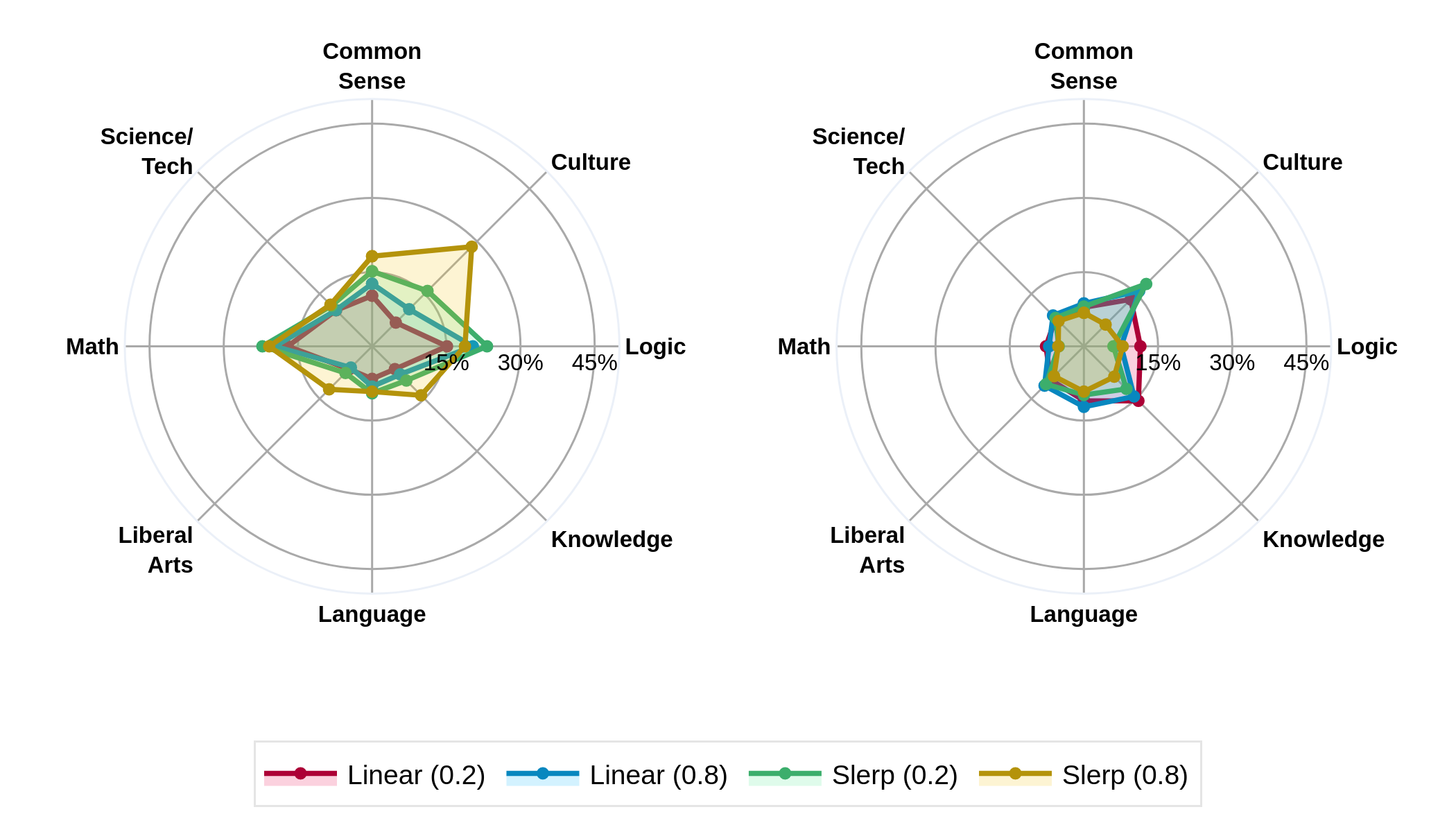}
  \caption{\textbf{Forgetting (left) and Backward Transfer (right) of Qwen 2.5 Base merged with Qwen 2.5 Coder (7B) relative to Qwen 2.5 Coder on all Benchmarks}. Moderate-to-large forgetting occurs with low-to-moderate backward transfer.}
\label{fig:merge-coder-after}
\end{figure}
\clearpage

\subsection*{Moderate success case: OpenThinker}

\begin{figure}[htbp]
  \centering
  \includegraphics[width=\textwidth,keepaspectratio]{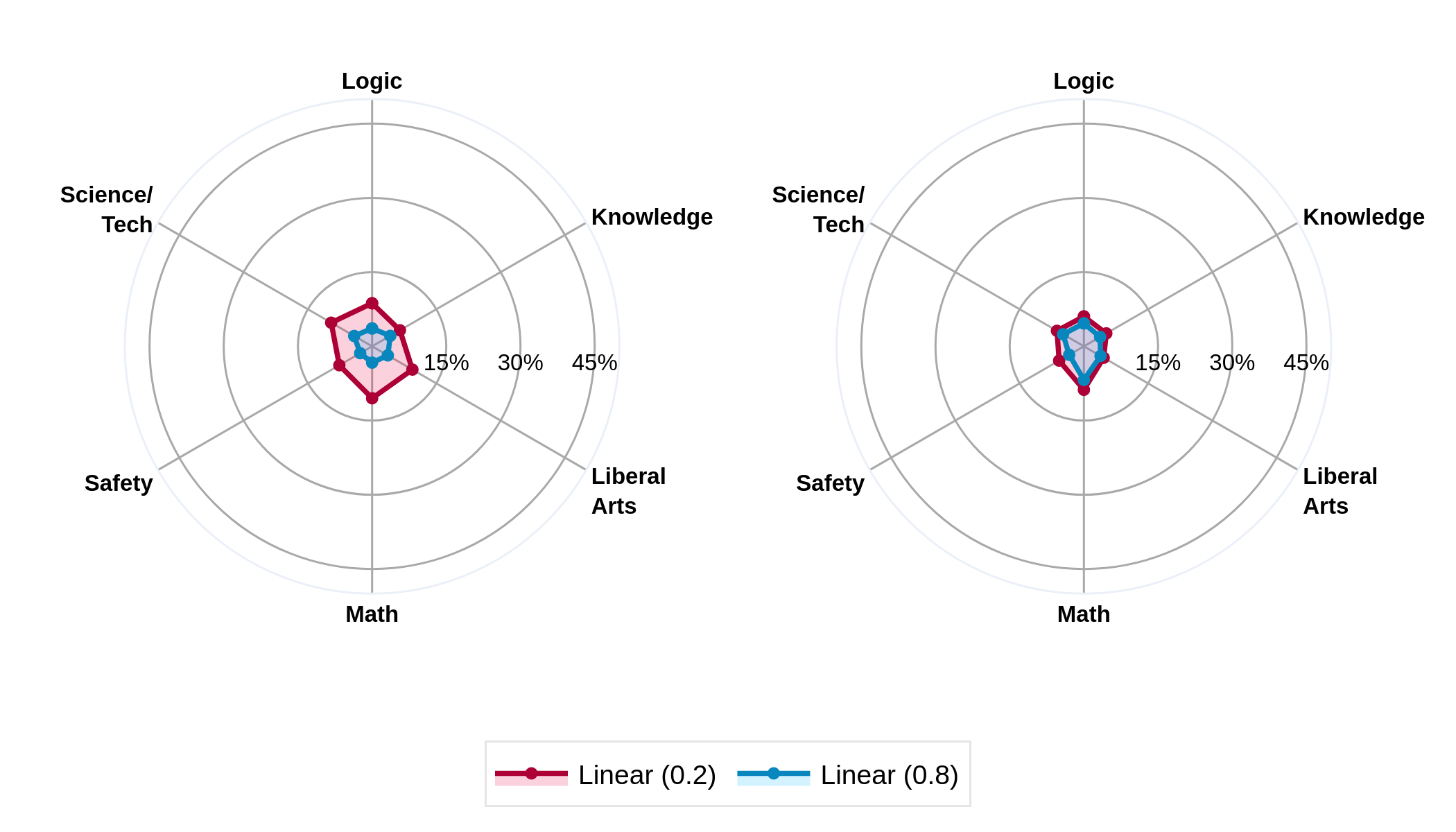}
  \caption{\textbf{Forgetting (left) and Backward Transfer (right) of Qwen 2.5 Instruct merged with OpenThinker Merge (7B)  relative to Qwen 2.5 Instruct on MMLU}. We see a marginal overall performance improvement in the case of Linear (0.8).}
\label{fig:merge-openthinker-before}
\end{figure}

\begin{figure}[htbp]
  \centering
  \includegraphics[width=\textwidth,keepaspectratio]{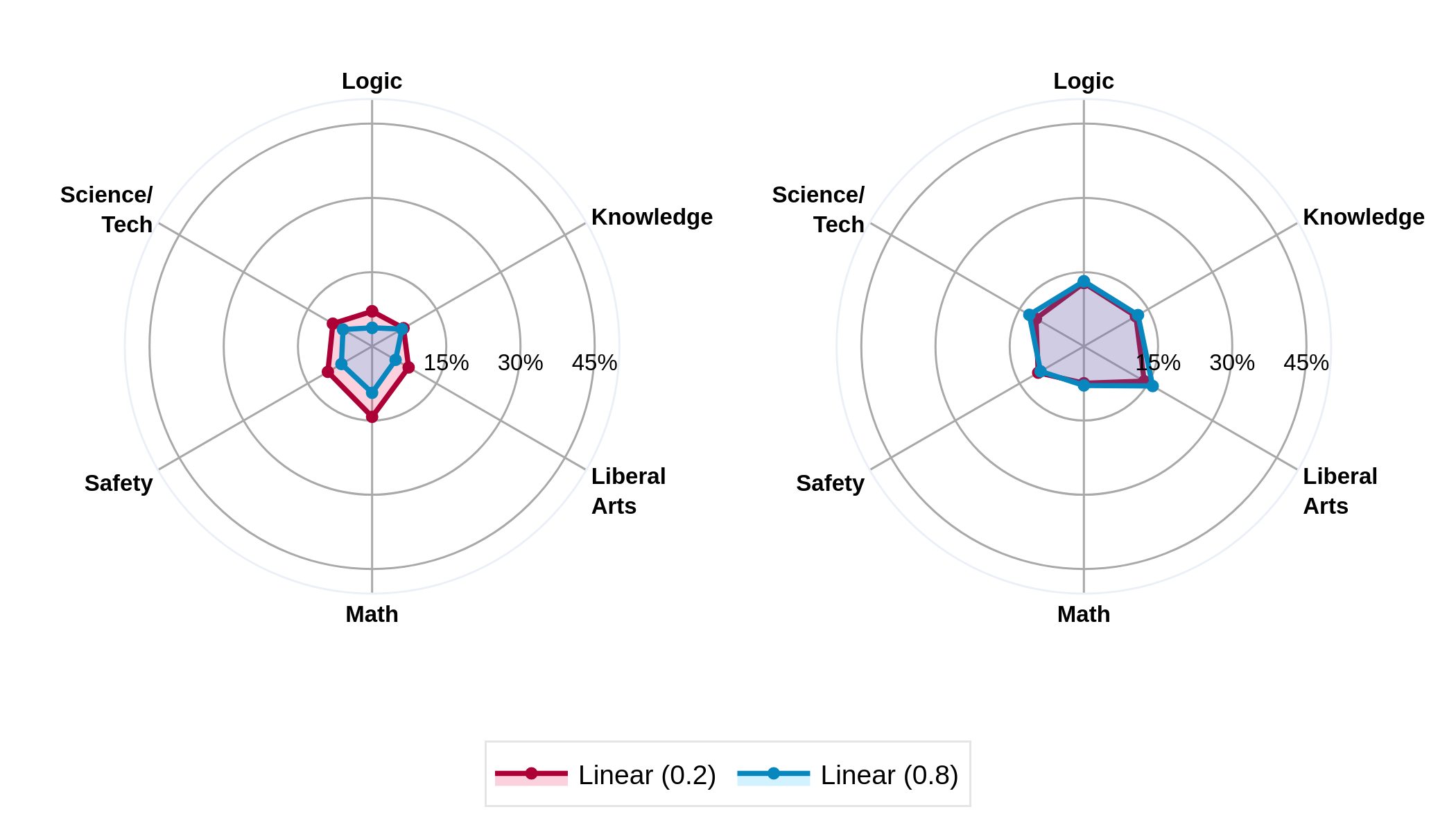}
  \caption{\textbf{Forgetting (left) and Backward Transfer (right)  of Qwen 2.5 Instruct merged with OpenThinker (7B) relative to OpenThinker on MMLU}. We see a marginal overall performance improvement in both cases.}
\label{fig:merge-openthinker-after}
\end{figure}
\clearpage

\section{Quantifying Forgetting Accurately (Tables for referencing plots)\protect\footnotemark[3]}
\footnotetext[3]{For brevity we shorten Qwen 2.5 to Q2.5 as well as the associated models (e.g. Qwen 2.5 Instruct to Q2.5 Inst.)}

Forgetting/backward transfer tables are listed with forgetting as the first number in each entry, standard deviation after the "$\pm$", and maximum possible forgetting/backward transfer resp. in brackets.

\subsection{Instruction Tuning}

\begin{table*}[htbp]
\centering
\caption{Instruction Tuning: Forgetting (Part 1 of 3)}
\label{tab:InstructionTuningAll_forgetting_part1}
\small
\begin{tabular}{lrrrr}
\toprule
Category & Q2.5 Inst. (3B) & Q2.5 Inst. (7B) & Q2.5 Inst. (14B) & Q2.5 Inst. (32B) \\
\midrule
Common Sense & 7.0 {\scriptsize$\pm$0.3} (64.2) & 5.4 {\scriptsize$\pm$0.1} (60.8) & 3.9 {\scriptsize$\pm$0.6} (75.8) & 3.7 {\scriptsize$\pm$0.5} (79.5) \\
Culture & 11.7 {\scriptsize$\pm$0.9} (55.8) & 16.5 {\scriptsize$\pm$2.8} (64.8) & 14.0 {\scriptsize$\pm$0.1} (76.4) & 15.2 {\scriptsize$\pm$0.6} (78.9) \\
Logic & 10.9 {\scriptsize$\pm$0.5} (36.2) & 8.4 {\scriptsize$\pm$0.2} (52.5) & 5.5 {\scriptsize$\pm$0.4} (65.3) & 4.6 {\scriptsize$\pm$0.6} (74.4) \\
Knowledge/QA & 6.8 {\scriptsize$\pm$1.3} (47.3) & 15.0 {\scriptsize$\pm$1.0} (58.4) & 23.4 {\scriptsize$\pm$0.3} (76.9) & 15.3 {\scriptsize$\pm$1.9} (69.4) \\
Language & 8.7 {\scriptsize$\pm$0.8} (31.0) & 9.2 {\scriptsize$\pm$0.6} (45.1) & 9.5 {\scriptsize$\pm$0.8} (59.2) & 8.6 {\scriptsize$\pm$1.0} (60.2) \\
Liberal Arts & 8.7 {\scriptsize$\pm$0.7} (65.3) & 6.6 {\scriptsize$\pm$0.7} (74.2) & 5.3 {\scriptsize$\pm$0.4} (78.6) & 5.3 {\scriptsize$\pm$0.3} (81.9) \\
Math & 7.7 {\scriptsize$\pm$0.4} (35.4) & 4.2 {\scriptsize$\pm$0.4} (47.3) & 6.2 {\scriptsize$\pm$0.7} (57.9) & 4.5 {\scriptsize$\pm$1.5} (64.4) \\
Science/Tech & 6.5 {\scriptsize$\pm$0.3} (45.6) & 5.4 {\scriptsize$\pm$0.4} (56.5) & 4.5 {\scriptsize$\pm$0.5} (65.2) & 4.4 {\scriptsize$\pm$0.3} (69.7) \\
\midrule
\textbf{Total} & 8.5 {\scriptsize$\pm$0.1} (50.2) & 8.2 {\scriptsize$\pm$0.3} (58.6) & 8.0 {\scriptsize$\pm$0.3} (69.7) & 6.7 {\scriptsize$\pm$0.5} (72.3) \\
\bottomrule
\end{tabular}
\end{table*}

\begin{table*}[htbp]
\centering
\caption{Instruction Tuning: Forgetting (Part 2 of 3)}
\label{tab:InstructionTuningAll_forgetting_part2}
\small
\begin{tabular}{lrrrr}
\toprule
Category & Q2.5 Coder Inst. (3B) & Q2.5 Coder Inst. (7B) & Q2.5 Coder Inst. (14B) & Q2.5 Coder Inst. (32B) \\
\midrule
Common Sense & 11.7 {\scriptsize$\pm$1.0} (59.5) & 8.1 {\scriptsize$\pm$0.3} (67.2) & 6.1 {\scriptsize$\pm$0.2} (70.7) & 4.6 {\scriptsize$\pm$0.4} (77.9) \\
Culture & 15.0 {\scriptsize$\pm$2.6} (45.7) & 19.0 {\scriptsize$\pm$1.9} (60.8) & 16.6 {\scriptsize$\pm$1.6} (66.9) & 19.1 {\scriptsize$\pm$1.1} (73.8) \\
Logic & 17.2 {\scriptsize$\pm$0.6} (41.2) & 14.1 {\scriptsize$\pm$0.2} (51.9) & 6.9 {\scriptsize$\pm$0.2} (64.0) & 5.8 {\scriptsize$\pm$0.3} (69.8) \\
Knowledge/QA & 12.6 {\scriptsize$\pm$0.1} (48.0) & 14.4 {\scriptsize$\pm$0.3} (56.4) & 14.3 {\scriptsize$\pm$0.3} (64.3) & 17.6 {\scriptsize$\pm$1.0} (77.7) \\
Language & 15.1 {\scriptsize$\pm$0.7} (36.4) & 13.7 {\scriptsize$\pm$1.0} (43.2) & 10.3 {\scriptsize$\pm$0.8} (52.7) & 8.6 {\scriptsize$\pm$0.5} (59.2) \\
Liberal Arts & 13.9 {\scriptsize$\pm$0.6} (59.2) & 9.6 {\scriptsize$\pm$0.0} (67.6) & 7.8 {\scriptsize$\pm$0.4} (74.6) & 6.7 {\scriptsize$\pm$0.2} (77.7) \\
Math & 8.9 {\scriptsize$\pm$0.9} (32.8) & 6.8 {\scriptsize$\pm$0.2} (40.8) & 6.2 {\scriptsize$\pm$0.4} (54.4) & 4.9 {\scriptsize$\pm$0.6} (58.6) \\
Science/Tech & 9.4 {\scriptsize$\pm$0.7} (40.5) & 8.4 {\scriptsize$\pm$0.5} (52.0) & 7.0 {\scriptsize$\pm$0.2} (59.9) & 5.6 {\scriptsize$\pm$0.5} (65.1) \\
\midrule
\textbf{Total} & 13.0 {\scriptsize$\pm$0.2} (48.2) & 10.9 {\scriptsize$\pm$0.3} (57.6) & 8.9 {\scriptsize$\pm$0.3} (66.2) & 8.4 {\scriptsize$\pm$0.1} (72.2) \\
\bottomrule
\end{tabular}
\end{table*}

\begin{table*}[htbp]
\centering
\caption{Instruction Tuning: Forgetting (Part 3 of 3)}
\label{tab:InstructionTuningAll_forgetting_part3}
\small
\begin{tabular}{lr}
\toprule
Category & Llama 3.1 Inst. (8B) \\
\midrule
Common Sense & 6.9 {\scriptsize$\pm$0.4} (64.5) \\
Culture & 25.3 {\scriptsize$\pm$2.2} (79.1) \\
Logic & 10.9 {\scriptsize$\pm$0.5} (42.5) \\
Knowledge/QA & 20.6 {\scriptsize$\pm$0.8} (60.8) \\
Language & 10.4 {\scriptsize$\pm$0.7} (39.9) \\
Liberal Arts & 7.6 {\scriptsize$\pm$0.6} (64.6) \\
Math & 7.3 {\scriptsize$\pm$0.9} (30.5) \\
Science/Tech & 5.9 {\scriptsize$\pm$0.9} (45.5) \\
\midrule
\textbf{Total} & 10.8 {\scriptsize$\pm$0.3} (54.5) \\
\bottomrule
\end{tabular}
\end{table*}

\begin{table*}[htbp]
\centering
\caption{Instruction Tuning: Backward Transfer (Part 1 of 3)}
\label{tab:InstructionTuningAll_btransfer_part1}
\small
\begin{tabular}{lrrrr}
\toprule
Category & Q2.5 Inst. (3B) & Q2.5 Inst. (7B) & Q2.5 Inst. (14B) & Q2.5 Inst. (32B) \\
\midrule
Common Sense & 10.8 {\scriptsize$\pm$0.4} (69.2) & 17.5 {\scriptsize$\pm$0.3} (76.9) & 8.7 {\scriptsize$\pm$0.1} (82.1) & 7.5 {\scriptsize$\pm$0.3} (84.6) \\
Culture & 7.9 {\scriptsize$\pm$2.4} (49.2) & 5.2 {\scriptsize$\pm$1.9} (45.3) & 5.3 {\scriptsize$\pm$0.3} (63.7) & 3.0 {\scriptsize$\pm$1.1} (62.1) \\
Logic & 10.2 {\scriptsize$\pm$0.7} (33.6) & 14.6 {\scriptsize$\pm$0.5} (60.4) & 13.3 {\scriptsize$\pm$0.5} (75.6) & 9.0 {\scriptsize$\pm$0.3} (80.4) \\
Knowledge/QA & 13.3 {\scriptsize$\pm$2.8} (55.8) & 7.5 {\scriptsize$\pm$2.0} (52.5) & 5.1 {\scriptsize$\pm$1.0} (57.9) & 7.7 {\scriptsize$\pm$1.7} (61.3) \\
Language & 7.7 {\scriptsize$\pm$0.5} (29.8) & 8.3 {\scriptsize$\pm$0.4} (41.4) & 7.2 {\scriptsize$\pm$0.0} (54.0) & 8.3 {\scriptsize$\pm$2.4} (57.0) \\
Liberal Arts & 7.2 {\scriptsize$\pm$1.4} (63.3) & 5.7 {\scriptsize$\pm$0.6} (73.0) & 5.6 {\scriptsize$\pm$0.9} (78.9) & 4.6 {\scriptsize$\pm$1.1} (80.9) \\
Math & 18.9 {\scriptsize$\pm$0.9} (51.0) & 19.0 {\scriptsize$\pm$0.8} (67.3) & 17.8 {\scriptsize$\pm$1.4} (73.8) & 15.9 {\scriptsize$\pm$1.9} (80.0) \\
Science/Tech & 11.4 {\scriptsize$\pm$0.8} (52.1) & 11.9 {\scriptsize$\pm$0.9} (65.2) & 10.4 {\scriptsize$\pm$1.0} (73.0) & 9.5 {\scriptsize$\pm$0.8} (76.5) \\
\midrule
\textbf{Total} & 10.5 {\scriptsize$\pm$0.5} (52.7) & 11.0 {\scriptsize$\pm$0.5} (62.1) & 8.9 {\scriptsize$\pm$0.4} (71.3) & 8.2 {\scriptsize$\pm$0.7} (74.3) \\
\bottomrule
\end{tabular}
\end{table*}

\begin{table*}[htbp]
\centering
\caption{Instruction Tuning: Backward Transfer (Part 2 of 3)}
\label{tab:InstructionTuningAll_btransfer_part2}
\small
\begin{tabular}{lrrrr}
\toprule
Category & Q2.5 Coder Inst. (3B) & Q2.5 Coder Inst. (7B) & Q2.5 Coder Inst. (14B) & Q2.5 Coder Inst. (32B) \\
\midrule
Common Sense & 10.9 {\scriptsize$\pm$0.1} (58.3) & 10.6 {\scriptsize$\pm$1.0} (70.5) & 10.2 {\scriptsize$\pm$0.8} (76.2) & 6.9 {\scriptsize$\pm$0.1} (80.9) \\
Culture & 4.8 {\scriptsize$\pm$0.9} (31.0) & 4.2 {\scriptsize$\pm$0.7} (37.8) & 2.8 {\scriptsize$\pm$0.5} (47.0) & 2.7 {\scriptsize$\pm$0.7} (50.0) \\
Logic & 6.6 {\scriptsize$\pm$0.3} (25.3) & 11.7 {\scriptsize$\pm$0.7} (47.2) & 11.5 {\scriptsize$\pm$0.3} (70.6) & 10.8 {\scriptsize$\pm$0.1} (76.4) \\
Knowledge/QA & 8.3 {\scriptsize$\pm$1.2} (43.4) & 9.4 {\scriptsize$\pm$1.9} (52.5) & 10.2 {\scriptsize$\pm$1.3} (60.6) & 4.8 {\scriptsize$\pm$1.5} (64.2) \\
Language & 4.7 {\scriptsize$\pm$0.4} (19.2) & 5.6 {\scriptsize$\pm$1.2} (29.4) & 7.9 {\scriptsize$\pm$0.2} (48.0) & 6.5 {\scriptsize$\pm$0.7} (55.2) \\
Liberal Arts & 5.8 {\scriptsize$\pm$1.1} (48.4) & 6.4 {\scriptsize$\pm$0.7} (63.3) & 5.5 {\scriptsize$\pm$1.1} (71.6) & 5.1 {\scriptsize$\pm$0.8} (75.6) \\
Math & 14.4 {\scriptsize$\pm$1.8} (41.0) & 21.7 {\scriptsize$\pm$1.4} (61.2) & 17.5 {\scriptsize$\pm$1.9} (69.6) & 19.1 {\scriptsize$\pm$1.5} (77.8) \\
Science/Tech & 8.7 {\scriptsize$\pm$1.6} (39.5) & 10.2 {\scriptsize$\pm$0.5} (54.3) & 10.4 {\scriptsize$\pm$1.0} (64.4) & 9.8 {\scriptsize$\pm$0.4} (70.6) \\
\midrule
\textbf{Total} & 7.4 {\scriptsize$\pm$0.2} (40.4) & 9.3 {\scriptsize$\pm$0.6} (54.9) & 8.6 {\scriptsize$\pm$0.3} (65.8) & 7.6 {\scriptsize$\pm$0.4} (71.2) \\
\bottomrule
\end{tabular}
\end{table*}

\begin{table*}[htbp]
\centering
\caption{Instruction Tuning: Backward Transfer (Part 3 of 3)}
\label{tab:InstructionTuningAll_btransfer_part3}
\small
\begin{tabular}{lr}
\toprule
Category & Llama 3.1 Inst. (8B) \\
\midrule
Common Sense & 11.2 {\scriptsize$\pm$0.3} (70.3) \\
Culture & 3.9 {\scriptsize$\pm$0.2} (46.2) \\
Logic & 17.0 {\scriptsize$\pm$0.9} (49.6) \\
Knowledge/QA & 10.8 {\scriptsize$\pm$2.3} (52.9) \\
Language & 8.9 {\scriptsize$\pm$1.4} (36.3) \\
Liberal Arts & 9.8 {\scriptsize$\pm$1.7} (67.5) \\
Math & 19.3 {\scriptsize$\pm$0.7} (46.9) \\
Science/Tech & 15.1 {\scriptsize$\pm$1.3} (57.8) \\
\midrule
\textbf{Total} & 11.4 {\scriptsize$\pm$1.0} (55.1) \\
\bottomrule
\end{tabular}
\end{table*}

\clearpage

\subsection{Domain-Continual Pretraining}

\begin{table*}[htbp]
\centering
\caption{Domain-Continual Pretraining: Forgetting (Part 1 of 2)}
\label{tab:TaskTrainingAll_forgetting_part1}
\small
\begin{tabular}{lrrrr}
\toprule
Category & Q2.5 Coder (3B) & Q2.5 Coder (7B) & Q2.5 Coder (14B) & Q2.5 Coder (32B) \\
\midrule
Common Sense & 11.9 {\scriptsize$\pm$0.5} (64.2) & 9.0 {\scriptsize$\pm$0.4} (60.8) & 10.4 {\scriptsize$\pm$0.6} (75.8) & 7.5 {\scriptsize$\pm$0.6} (79.5) \\
Culture & 11.7 {\scriptsize$\pm$0.7} (55.8) & 10.9 {\scriptsize$\pm$0.4} (64.8) & 10.8 {\scriptsize$\pm$0.7} (76.4) & 8.8 {\scriptsize$\pm$0.8} (78.9) \\
Logic & 5.7 {\scriptsize$\pm$0.2} (36.2) & 9.5 {\scriptsize$\pm$0.2} (52.5) & 7.6 {\scriptsize$\pm$0.3} (65.3) & 8.4 {\scriptsize$\pm$0.1} (74.4) \\
Knowledge/QA & 5.6 {\scriptsize$\pm$0.5} (47.3) & 6.0 {\scriptsize$\pm$0.8} (58.4) & 13.7 {\scriptsize$\pm$0.2} (76.9) & 3.8 {\scriptsize$\pm$0.5} (69.4) \\
Language & 5.9 {\scriptsize$\pm$0.6} (31.0) & 8.4 {\scriptsize$\pm$1.5} (45.1) & 8.6 {\scriptsize$\pm$0.9} (59.2) & 7.4 {\scriptsize$\pm$1.3} (60.2) \\
Liberal Arts & 6.4 {\scriptsize$\pm$0.5} (65.3) & 7.0 {\scriptsize$\pm$0.3} (74.2) & 5.1 {\scriptsize$\pm$0.4} (78.6) & 5.2 {\scriptsize$\pm$0.3} (81.9) \\
Math & 3.8 {\scriptsize$\pm$0.9} (35.4) & 7.4 {\scriptsize$\pm$1.0} (47.3) & 6.2 {\scriptsize$\pm$0.2} (57.9) & 7.8 {\scriptsize$\pm$0.5} (64.4) \\
Science/Tech & 4.0 {\scriptsize$\pm$0.5} (45.6) & 5.9 {\scriptsize$\pm$0.3} (56.5) & 6.4 {\scriptsize$\pm$0.5} (65.2) & 5.7 {\scriptsize$\pm$0.5} (69.7) \\
\midrule
\textbf{Total} & 6.8 {\scriptsize$\pm$0.0} (50.9) & 7.6 {\scriptsize$\pm$0.2} (60.4) & 8.0 {\scriptsize$\pm$0.1} (71.8) & 6.4 {\scriptsize$\pm$0.2} (74.6) \\
\bottomrule
\end{tabular}
\end{table*}

\begin{table*}[htbp]
\centering
\caption{Domain-Continual Pretraining: Forgetting (Part 2 of 2)}
\label{tab:TaskTrainingAll_forgetting_part2}
\small
\begin{tabular}{lr}
\toprule
Category & Q2.5 Math (7B) \\
\midrule
Common Sense & 13.6 {\scriptsize$\pm$0.8} (60.8) \\
Culture & 17.8 {\scriptsize$\pm$0.3} (64.8) \\
Logic & 9.8 {\scriptsize$\pm$0.4} (52.5) \\
Knowledge/QA & 9.8 {\scriptsize$\pm$0.5} (58.4) \\
Language & 11.3 {\scriptsize$\pm$0.9} (45.1) \\
Liberal Arts & 20.0 {\scriptsize$\pm$1.3} (74.2) \\
Math & 7.5 {\scriptsize$\pm$1.2} (47.3) \\
Science/Tech & 14.4 {\scriptsize$\pm$0.5} (56.5) \\
\midrule
\textbf{Total} & 12.9 {\scriptsize$\pm$0.4} (60.4) \\
\bottomrule
\end{tabular}
\end{table*}

\begin{table*}[htbp]
\centering
\caption{Domain-Continual Pretraining: Backward Transfer (Part 1 of 2)}
\label{tab:TaskTrainingAll_btransfer_part1}
\small
\begin{tabular}{lrrrr}
\toprule
Category & Q2.5 Coder (3B) & Q2.5 Coder (7B) & Q2.5 Coder (14B) & Q2.5 Coder (32B) \\
\midrule
Common Sense & 8.3 {\scriptsize$\pm$0.4} (59.5) & 13.8 {\scriptsize$\pm$0.3} (67.2) & 6.6 {\scriptsize$\pm$0.3} (70.7) & 6.3 {\scriptsize$\pm$0.5} (77.9) \\
Culture & 6.3 {\scriptsize$\pm$2.1} (45.7) & 10.0 {\scriptsize$\pm$1.3} (60.8) & 5.8 {\scriptsize$\pm$0.4} (66.9) & 5.6 {\scriptsize$\pm$0.2} (73.8) \\
Logic & 9.4 {\scriptsize$\pm$0.5} (41.2) & 9.7 {\scriptsize$\pm$0.2} (51.9) & 7.7 {\scriptsize$\pm$0.8} (64.0) & 5.3 {\scriptsize$\pm$0.4} (69.8) \\
Knowledge/QA & 6.6 {\scriptsize$\pm$1.1} (48.0) & 5.4 {\scriptsize$\pm$0.5} (56.4) & 2.7 {\scriptsize$\pm$0.3} (64.3) & 12.4 {\scriptsize$\pm$0.7} (77.7) \\
Language & 8.9 {\scriptsize$\pm$0.9} (36.4) & 7.6 {\scriptsize$\pm$0.8} (43.2) & 5.3 {\scriptsize$\pm$0.5} (52.7) & 7.8 {\scriptsize$\pm$2.2} (59.2) \\
Liberal Arts & 2.0 {\scriptsize$\pm$0.2} (59.2) & 2.2 {\scriptsize$\pm$0.3} (67.6) & 2.3 {\scriptsize$\pm$0.3} (74.6) & 2.1 {\scriptsize$\pm$0.1} (77.7) \\
Math & 2.4 {\scriptsize$\pm$0.2} (32.8) & 3.2 {\scriptsize$\pm$0.3} (40.8) & 3.4 {\scriptsize$\pm$1.2} (54.4) & 3.4 {\scriptsize$\pm$1.0} (58.6) \\
Science/Tech & 1.0 {\scriptsize$\pm$0.3} (40.5) & 2.8 {\scriptsize$\pm$0.0} (52.0) & 2.6 {\scriptsize$\pm$0.1} (59.9) & 2.4 {\scriptsize$\pm$0.2} (65.1) \\
\midrule
\textbf{Total} & 5.1 {\scriptsize$\pm$0.2} (48.2) & 6.2 {\scriptsize$\pm$0.1} (57.6) & 4.2 {\scriptsize$\pm$0.2} (66.2) & 5.1 {\scriptsize$\pm$0.2} (72.2) \\
\bottomrule
\end{tabular}
\end{table*}

\begin{table*}[htbp]
\centering
\caption{Domain-Continual Pretraining: Backward Transfer (Part 2 of 2)}
\label{tab:TaskTrainingAll_btransfer_part2}
\small
\begin{tabular}{lr}
\toprule
Category & Q2.5 Math (7B) \\
\midrule
Common Sense & 12.7 {\scriptsize$\pm$0.5} (59.6) \\
Culture & 6.9 {\scriptsize$\pm$0.8} (46.3) \\
Logic & 11.9 {\scriptsize$\pm$0.2} (53.9) \\
Knowledge/QA & 9.9 {\scriptsize$\pm$1.8} (55.3) \\
Language & 6.1 {\scriptsize$\pm$0.2} (36.3) \\
Liberal Arts & 1.6 {\scriptsize$\pm$0.3} (49.5) \\
Math & 4.3 {\scriptsize$\pm$0.5} (43.1) \\
Science/Tech & 3.2 {\scriptsize$\pm$0.7} (40.7) \\
\midrule
\textbf{Total} & 6.3 {\scriptsize$\pm$0.2} (50.1) \\
\bottomrule
\end{tabular}
\end{table*}

\clearpage

\subsection{Trained from Base}\label{sec:reasoning_from_base}

\begin{table*}[htbp]
\centering
\caption{Trained from Base: Forgetting (Part 1 of 2)}
\label{tab:TrainedfromBaseAll_forgetting_part1}
\small
\begin{tabular}{lrrrr}
\toprule
Category & Q2.5 Math Inst. (7B) & QwQ (32B) & R1 Distill Qwen (7B) & R1 Distill Llama (8B) \\
\midrule
Common Sense & 26.0 {\scriptsize$\pm$0.3} (59.6) & 3.2 {\scriptsize$\pm$0.4} (79.5) & 8.9 {\scriptsize$\pm$0.3} (59.6) & 8.1 {\scriptsize$\pm$0.2} (64.5) \\
Culture & 27.0 {\scriptsize$\pm$2.4} (46.3) & 15.7 {\scriptsize$\pm$0.6} (78.9) & 19.3 {\scriptsize$\pm$2.1} (46.3) & 29.0 {\scriptsize$\pm$1.0} (79.1) \\
Logic & 16.8 {\scriptsize$\pm$0.4} (53.9) & 2.8 {\scriptsize$\pm$0.2} (74.4) & 4.4 {\scriptsize$\pm$0.4} (53.9) & 4.8 {\scriptsize$\pm$0.2} (42.5) \\
Knowledge/QA & 25.4 {\scriptsize$\pm$2.0} (55.3) & 13.2 {\scriptsize$\pm$2.1} (69.4) & 20.1 {\scriptsize$\pm$0.7} (55.3) & 11.7 {\scriptsize$\pm$0.7} (60.8) \\
Language & 18.2 {\scriptsize$\pm$0.9} (36.3) & 7.9 {\scriptsize$\pm$1.2} (60.2) & 9.4 {\scriptsize$\pm$0.8} (36.3) & 10.4 {\scriptsize$\pm$0.4} (39.9) \\
Liberal Arts & 23.7 {\scriptsize$\pm$0.7} (49.5) & 3.1 {\scriptsize$\pm$0.2} (81.9) & 7.6 {\scriptsize$\pm$0.4} (49.5) & 9.4 {\scriptsize$\pm$0.9} (64.6) \\
Math & 13.1 {\scriptsize$\pm$0.3} (43.1) & 1.5 {\scriptsize$\pm$0.3} (64.4) & 2.2 {\scriptsize$\pm$0.4} (43.1) & 3.3 {\scriptsize$\pm$0.3} (30.5) \\
Science/Tech & 17.1 {\scriptsize$\pm$0.9} (40.7) & 2.2 {\scriptsize$\pm$0.2} (69.7) & 4.6 {\scriptsize$\pm$0.2} (40.7) & 6.4 {\scriptsize$\pm$0.4} (45.5) \\
\midrule
\textbf{Total} & 21.4 {\scriptsize$\pm$0.5} (50.1) & 5.4 {\scriptsize$\pm$0.3} (72.3) & 8.7 {\scriptsize$\pm$0.1} (50.1) & 9.3 {\scriptsize$\pm$0.3} (54.5) \\
\bottomrule
\end{tabular}
\end{table*}

\begin{table*}[htbp]
\centering
\caption{Trained from Base: Forgetting (Part 2 of 2)}
\label{tab:TrainedfromBaseAll_forgetting_part2}
\small
\begin{tabular}{lr}
\toprule
Category & R1 Distill Qwen (32B) \\
\midrule
Common Sense & 3.9 {\scriptsize$\pm$0.3} (79.5) \\
Culture & 18.8 {\scriptsize$\pm$0.7} (78.9) \\
Logic & 2.3 {\scriptsize$\pm$0.3} (74.4) \\
Knowledge/QA & 15.3 {\scriptsize$\pm$1.3} (69.4) \\
Language & 6.9 {\scriptsize$\pm$1.4} (60.2) \\
Liberal Arts & 3.6 {\scriptsize$\pm$0.3} (81.9) \\
Math & 1.8 {\scriptsize$\pm$0.5} (64.4) \\
Science/Tech & 2.4 {\scriptsize$\pm$0.3} (69.7) \\
\midrule
\textbf{Total} & 6.0 {\scriptsize$\pm$0.2} (72.3) \\
\bottomrule
\end{tabular}
\end{table*}

\begin{table*}[htbp]
\centering
\caption{Trained from Base: Backward Transfer (Part 1 of 2)}
\label{tab:TrainedfromBaseAll_btransfer_part1}
\small
\begin{tabular}{lrrrr}
\toprule
Category & Q2.5 Math Inst. (7B) & QwQ (32B) & R1 Distill Qwen (7B) & R1 Distill Llama (8B) \\
\midrule
Common Sense & 6.0 {\scriptsize$\pm$0.6} (32.9) & 8.9 {\scriptsize$\pm$0.5} (87.0) & 11.4 {\scriptsize$\pm$0.1} (63.0) & 12.6 {\scriptsize$\pm$0.8} (70.5) \\
Culture & 3.4 {\scriptsize$\pm$1.3} (9.0) & 2.9 {\scriptsize$\pm$1.0} (59.8) & 5.6 {\scriptsize$\pm$1.4} (23.1) & 2.8 {\scriptsize$\pm$0.4} (36.5) \\
Logic & 8.4 {\scriptsize$\pm$0.4} (42.8) & 11.8 {\scriptsize$\pm$0.5} (86.7) & 19.7 {\scriptsize$\pm$0.1} (74.9) & 28.9 {\scriptsize$\pm$0.1} (74.6) \\
Knowledge/QA & 4.1 {\scriptsize$\pm$1.4} (31.2) & 9.0 {\scriptsize$\pm$1.6} (65.2) & 9.3 {\scriptsize$\pm$2.4} (45.6) & 10.8 {\scriptsize$\pm$1.8} (61.8) \\
Language & 9.6 {\scriptsize$\pm$0.6} (20.9) & 9.7 {\scriptsize$\pm$2.8} (60.6) & 8.1 {\scriptsize$\pm$0.8} (33.7) & 12.1 {\scriptsize$\pm$1.0} (42.2) \\
Liberal Arts & 5.0 {\scriptsize$\pm$0.4} (24.5) & 6.3 {\scriptsize$\pm$1.0} (86.1) & 12.7 {\scriptsize$\pm$1.6} (56.2) & 10.3 {\scriptsize$\pm$1.5} (65.8) \\
Math & 15.1 {\scriptsize$\pm$1.6} (46.5) & 22.5 {\scriptsize$\pm$2.2} (92.2) & 34.3 {\scriptsize$\pm$2.4} (85.8) & 35.0 {\scriptsize$\pm$0.5} (72.8) \\
Science/Tech & 7.2 {\scriptsize$\pm$0.2} (27.4) & 13.8 {\scriptsize$\pm$1.0} (85.0) & 18.5 {\scriptsize$\pm$1.5} (59.3) & 19.3 {\scriptsize$\pm$1.2} (62.7) \\
\midrule
\textbf{Total} & 6.7 {\scriptsize$\pm$0.2} (30.1) & 10.3 {\scriptsize$\pm$0.6} (78.6) & 14.3 {\scriptsize$\pm$1.0} (57.4) & 15.4 {\scriptsize$\pm$0.6} (62.1) \\
\bottomrule
\end{tabular}
\end{table*}

\begin{table*}[htbp]
\centering
\caption{Trained from Base: Backward Transfer (Part 2 of 2)}
\label{tab:TrainedfromBaseAll_btransfer_part2}
\small
\begin{tabular}{lr}
\toprule
Category & R1 Distill Qwen (32B) \\
\midrule
Common Sense & 8.7 {\scriptsize$\pm$0.3} (85.9) \\
Culture & 2.5 {\scriptsize$\pm$0.6} (55.4) \\
Logic & 11.7 {\scriptsize$\pm$0.4} (87.2) \\
Knowledge/QA & 6.8 {\scriptsize$\pm$1.3} (60.6) \\
Language & 10.2 {\scriptsize$\pm$2.7} (62.9) \\
Liberal Arts & 6.1 {\scriptsize$\pm$1.0} (85.2) \\
Math & 22.1 {\scriptsize$\pm$2.5} (91.4) \\
Science/Tech & 13.3 {\scriptsize$\pm$1.0} (84.2) \\
\midrule
\textbf{Total} & 10.0 {\scriptsize$\pm$0.6} (77.5) \\
\bottomrule
\end{tabular}
\end{table*}

\clearpage

\subsection{Trained from Instruct - High Data Scenario}

\begin{table*}[htbp]
\centering
\caption{Trained from Instruct - High Data Scenario: Forgetting (Part 1 of 2)}
\label{tab:TrainedfromInstructHighDataScenarioAll_forgetting_part1}
\small
\begin{tabular}{lrrrr}
\toprule
Category & INTELLECT-2 (32B) & Open Math 2 (8B) & OpenThinker (7B) & OpenThinker2 (32B) \\
\midrule
Common Sense & 1.9 {\scriptsize$\pm$0.3} (87.0) & 28.5 {\scriptsize$\pm$0.9} (64.5) & 18.1 {\scriptsize$\pm$1.3} (76.9) & 3.9 {\scriptsize$\pm$0.3} (84.6) \\
Culture & 0.7 {\scriptsize$\pm$0.6} (60.2) & 49.9 {\scriptsize$\pm$1.9} (79.1) & 13.3 {\scriptsize$\pm$4.6} (45.3) & 7.6 {\scriptsize$\pm$2.1} (62.1) \\
Logic & 0.6 {\scriptsize$\pm$0.1} (87.4) & 24.2 {\scriptsize$\pm$0.5} (42.5) & 7.4 {\scriptsize$\pm$0.2} (60.4) & 1.0 {\scriptsize$\pm$0.2} (80.4) \\
Knowledge/QA & 2.5 {\scriptsize$\pm$0.6} (65.9) & 33.8 {\scriptsize$\pm$1.4} (60.8) & 13.4 {\scriptsize$\pm$1.1} (52.5) & 2.8 {\scriptsize$\pm$0.5} (61.3) \\
Language & 1.9 {\scriptsize$\pm$0.3} (62.7) & 20.9 {\scriptsize$\pm$1.2} (39.9) & 10.7 {\scriptsize$\pm$1.2} (41.4) & 4.2 {\scriptsize$\pm$0.4} (57.0) \\
Liberal Arts & 1.2 {\scriptsize$\pm$0.2} (86.1) & 32.1 {\scriptsize$\pm$2.3} (64.6) & 15.8 {\scriptsize$\pm$1.0} (73.0) & 2.3 {\scriptsize$\pm$0.1} (80.9) \\
Math & 0.9 {\scriptsize$\pm$0.1} (91.9) & 17.1 {\scriptsize$\pm$1.4} (30.5) & 8.3 {\scriptsize$\pm$0.8} (67.3) & 1.0 {\scriptsize$\pm$0.3} (80.0) \\
Safety/Truth & 0.9 {\scriptsize$\pm$0.2} (66.8) & 19.5 {\scriptsize$\pm$0.7} (36.3) & 8.6 {\scriptsize$\pm$0.5} (50.0) & 3.3 {\scriptsize$\pm$0.2} (64.5) \\
Science/Tech & 1.6 {\scriptsize$\pm$0.1} (85.0) & 24.8 {\scriptsize$\pm$1.8} (45.5) & 12.2 {\scriptsize$\pm$0.5} (65.2) & 2.1 {\scriptsize$\pm$0.2} (76.5) \\
\midrule
\textbf{Total} & 1.3 {\scriptsize$\pm$0.1} (79.0) & 28.8 {\scriptsize$\pm$0.8} (54.5) & 12.6 {\scriptsize$\pm$1.5} (62.0) & 2.9 {\scriptsize$\pm$0.2} (74.3) \\
\bottomrule
\end{tabular}
\end{table*}

\begin{table*}[htbp]
\centering
\caption{Trained from Instruct - High Data Scenario: Forgetting (Part 2 of 2)}
\label{tab:TrainedfromInstructHighDataScenarioAll_forgetting_part2}
\small
\begin{tabular}{lrrr}
\toprule
Category & OpenThinker3 (7B) & Nemotron Code Reasoning (7B) & Nemotron Code Reasoning (14B) \\
\midrule
Common Sense & 9.5 {\scriptsize$\pm$0.4} (76.9) & 22.1 {\scriptsize$\pm$0.1} (76.9) & 20.9 {\scriptsize$\pm$2.4} (79.0) \\
Culture & 16.4 {\scriptsize$\pm$4.5} (45.3) & 21.8 {\scriptsize$\pm$3.1} (45.3) & 20.9 {\scriptsize$\pm$0.7} (63.7) \\
Logic & 5.2 {\scriptsize$\pm$0.2} (60.4) & 14.3 {\scriptsize$\pm$0.4} (60.4) & 14.7 {\scriptsize$\pm$0.3} (75.6) \\
Knowledge/QA & 5.6 {\scriptsize$\pm$0.9} (52.5) & 11.9 {\scriptsize$\pm$1.6} (52.5) & 11.5 {\scriptsize$\pm$2.1} (57.9) \\
Language & 12.7 {\scriptsize$\pm$0.8} (41.4) & 21.2 {\scriptsize$\pm$1.4} (48.3) & 18.4 {\scriptsize$\pm$1.2} (62.4) \\
Liberal Arts & 9.8 {\scriptsize$\pm$0.6} (73.0) & 19.4 {\scriptsize$\pm$0.7} (73.0) & 12.8 {\scriptsize$\pm$1.0} (79.6) \\
Math & 4.1 {\scriptsize$\pm$0.1} (67.3) & 16.6 {\scriptsize$\pm$0.5} (67.3) & 13.3 {\scriptsize$\pm$0.4} (73.8) \\
Safety/Truth & 8.7 {\scriptsize$\pm$0.6} (50.0) & 14.2 {\scriptsize$\pm$0.1} (50.0) & 12.3 {\scriptsize$\pm$1.2} (63.0) \\
Science/Tech & 6.4 {\scriptsize$\pm$0.3} (65.2) & 18.2 {\scriptsize$\pm$0.4} (65.2) & 13.0 {\scriptsize$\pm$0.2} (73.0) \\
\midrule
\textbf{Total} & 8.4 {\scriptsize$\pm$0.5} (62.1) & 17.7 {\scriptsize$\pm$0.3} (62.4) & 14.9 {\scriptsize$\pm$0.0} (72.1) \\
\bottomrule
\end{tabular}
\end{table*}

\begin{table*}[htbp]
\centering
\caption{Trained from Instruct - High Data Scenario: Backward Transfer (Part 1 of 2)}
\label{tab:TrainedfromInstructHighDataScenarioAll_btransfer_part1}
\small
\begin{tabular}{lrrrr}
\toprule
Category & INTELLECT-2 (32B) & Open Math 2 (8B) & OpenThinker (7B) & OpenThinker2 (32B) \\
\midrule
Common Sense & 1.7 {\scriptsize$\pm$0.1} (86.7) & 5.7 {\scriptsize$\pm$0.4} (34.1) & 6.1 {\scriptsize$\pm$0.8} (60.9) & 4.2 {\scriptsize$\pm$0.5} (85.0) \\
Culture & 1.1 {\scriptsize$\pm$0.3} (60.8) & 0.5 {\scriptsize$\pm$0.4} (6.5) & 5.7 {\scriptsize$\pm$1.6} (33.6) & 2.9 {\scriptsize$\pm$0.3} (54.3) \\
Logic & 0.7 {\scriptsize$\pm$0.1} (87.4) & 6.0 {\scriptsize$\pm$0.4} (18.3) & 13.7 {\scriptsize$\pm$0.6} (67.4) & 6.2 {\scriptsize$\pm$0.4} (87.5) \\
Knowledge/QA & 3.4 {\scriptsize$\pm$0.9} (67.5) & 2.8 {\scriptsize$\pm$1.2} (25.3) & 7.0 {\scriptsize$\pm$0.7} (43.0) & 3.8 {\scriptsize$\pm$0.1} (62.3) \\
Language & 1.7 {\scriptsize$\pm$0.2} (62.7) & 7.2 {\scriptsize$\pm$0.8} (18.1) & 8.8 {\scriptsize$\pm$1.6} (39.1) & 6.4 {\scriptsize$\pm$0.7} (61.0) \\
Liberal Arts & 1.2 {\scriptsize$\pm$0.1} (86.0) & 2.9 {\scriptsize$\pm$0.2} (25.5) & 5.5 {\scriptsize$\pm$0.5} (59.2) & 5.6 {\scriptsize$\pm$0.3} (85.4) \\
Math & 0.9 {\scriptsize$\pm$0.2} (92.1) & 6.9 {\scriptsize$\pm$1.1} (15.0) & 11.1 {\scriptsize$\pm$1.0} (71.3) & 10.1 {\scriptsize$\pm$1.1} (91.7) \\
Safety/Truth & 1.3 {\scriptsize$\pm$0.5} (67.2) & 4.0 {\scriptsize$\pm$0.6} (15.9) & 7.2 {\scriptsize$\pm$1.2} (48.1) & 4.1 {\scriptsize$\pm$0.8} (65.5) \\
Science/Tech & 1.5 {\scriptsize$\pm$0.1} (85.0) & 3.6 {\scriptsize$\pm$0.9} (17.0) & 7.4 {\scriptsize$\pm$0.4} (58.7) & 7.6 {\scriptsize$\pm$0.3} (83.8) \\
\midrule
\textbf{Total} & 1.4 {\scriptsize$\pm$0.1} (79.2) & 4.2 {\scriptsize$\pm$0.2} (21.1) & 7.7 {\scriptsize$\pm$0.5} (55.1) & 5.3 {\scriptsize$\pm$0.2} (77.3) \\
\bottomrule
\end{tabular}
\end{table*}

\begin{table*}[htbp]
\centering
\caption{Trained from Instruct - High Data Scenario: Backward Transfer (Part 2 of 2)}
\label{tab:TrainedfromInstructHighDataScenarioAll_btransfer_part2}
\small
\begin{tabular}{lrrr}
\toprule
Category & OpenThinker3 (7B) & Nemotron Code Reasoning (7B) & Nemotron Code Reasoning (14B) \\
\midrule
Common Sense & 6.7 {\scriptsize$\pm$0.2} (73.2) & 5.2 {\scriptsize$\pm$0.4} (54.3) & 3.6 {\scriptsize$\pm$1.2} (56.0) \\
Culture & 4.0 {\scriptsize$\pm$1.0} (23.3) & 5.1 {\scriptsize$\pm$2.3} (19.2) & 5.2 {\scriptsize$\pm$0.5} (37.8) \\
Logic & 17.2 {\scriptsize$\pm$0.9} (76.3) & 13.7 {\scriptsize$\pm$0.7} (57.4) & 8.1 {\scriptsize$\pm$0.3} (65.4) \\
Knowledge/QA & 20.0 {\scriptsize$\pm$0.3} (66.6) & 5.0 {\scriptsize$\pm$0.5} (42.6) & 3.2 {\scriptsize$\pm$0.2} (46.6) \\
Language & 6.8 {\scriptsize$\pm$0.7} (33.6) & 4.9 {\scriptsize$\pm$0.8} (23.3) & 7.6 {\scriptsize$\pm$0.8} (46.3) \\
Liberal Arts & 5.7 {\scriptsize$\pm$0.2} (67.5) & 4.3 {\scriptsize$\pm$0.1} (52.8) & 4.2 {\scriptsize$\pm$0.5} (68.3) \\
Math & 14.7 {\scriptsize$\pm$0.3} (81.2) & 10.4 {\scriptsize$\pm$0.0} (56.4) & 10.7 {\scriptsize$\pm$0.8} (69.4) \\
Safety/Truth & 7.5 {\scriptsize$\pm$0.9} (48.5) & 8.6 {\scriptsize$\pm$1.2} (42.5) & 6.8 {\scriptsize$\pm$0.6} (55.6) \\
Science/Tech & 9.4 {\scriptsize$\pm$0.5} (69.2) & 5.2 {\scriptsize$\pm$0.0} (47.7) & 6.4 {\scriptsize$\pm$0.1} (64.3) \\
\midrule
\textbf{Total} & 9.7 {\scriptsize$\pm$0.2} (62.8) & 6.7 {\scriptsize$\pm$0.4} (46.5) & 5.8 {\scriptsize$\pm$0.1} (59.0) \\
\bottomrule
\end{tabular}
\end{table*}

\clearpage

\subsection{Trained from Instruct - Low Data Scenario}

\begin{table*}[htbp]
\centering
\caption{Trained from Instruct - Low Data Scenario: Forgetting (Part 1 of 2)}
\label{tab:TrainedfromInstructLowDataScenario_forgetting_part1}
\small
\begin{tabular}{lrrrr}
\toprule
Category & s1.1 (7B) & s1.1 (14B) & s1.1 (32B) & LIMO (32B) \\
\midrule
Common Sense & 8.1 {\scriptsize$\pm$0.4} (76.9) & 5.1 {\scriptsize$\pm$1.0} (82.1) & 4.7 {\scriptsize$\pm$0.5} (84.6) & 3.5 {\scriptsize$\pm$0.3} (84.6) \\
Culture & 12.1 {\scriptsize$\pm$1.2} (45.3) & 10.6 {\scriptsize$\pm$0.4} (63.7) & 7.5 {\scriptsize$\pm$0.9} (62.1) & 3.7 {\scriptsize$\pm$1.9} (62.1) \\
Logic & 7.2 {\scriptsize$\pm$0.1} (60.4) & 4.2 {\scriptsize$\pm$0.4} (75.6) & 2.8 {\scriptsize$\pm$0.2} (80.4) & 3.0 {\scriptsize$\pm$0.3} (80.4) \\
Knowledge/QA & 5.7 {\scriptsize$\pm$1.1} (52.5) & 2.3 {\scriptsize$\pm$0.1} (57.9) & 2.3 {\scriptsize$\pm$1.2} (61.3) & 2.0 {\scriptsize$\pm$0.4} (61.3) \\
Language & 8.6 {\scriptsize$\pm$0.8} (41.4) & 6.7 {\scriptsize$\pm$0.4} (54.0) & 4.9 {\scriptsize$\pm$0.1} (57.0) & 3.8 {\scriptsize$\pm$0.6} (57.0) \\
Liberal Arts & 7.4 {\scriptsize$\pm$0.5} (73.0) & 5.4 {\scriptsize$\pm$0.7} (78.9) & 3.9 {\scriptsize$\pm$0.1} (80.9) & 2.6 {\scriptsize$\pm$0.0} (80.9) \\
Math & 6.6 {\scriptsize$\pm$0.8} (67.3) & 5.8 {\scriptsize$\pm$0.9} (73.8) & 3.2 {\scriptsize$\pm$0.4} (80.0) & 1.3 {\scriptsize$\pm$0.1} (80.0) \\
Safety/Truth & 7.5 {\scriptsize$\pm$0.9} (50.0) & 5.3 {\scriptsize$\pm$0.3} (59.4) & 4.5 {\scriptsize$\pm$0.4} (64.5) & 2.9 {\scriptsize$\pm$0.3} (64.5) \\
Science/Tech & 7.2 {\scriptsize$\pm$0.4} (65.2) & 4.8 {\scriptsize$\pm$0.6} (73.0) & 3.3 {\scriptsize$\pm$0.4} (76.5) & 2.3 {\scriptsize$\pm$0.1} (76.5) \\
\midrule
\textbf{Total} & 7.7 {\scriptsize$\pm$0.2} (62.1) & 5.3 {\scriptsize$\pm$0.2} (71.3) & 3.9 {\scriptsize$\pm$0.1} (74.3) & 2.6 {\scriptsize$\pm$0.1} (74.3) \\
\bottomrule
\end{tabular}
\end{table*}

\begin{table*}[htbp]
\centering
\caption{Trained from Instruct - Low Data Scenario: Forgetting (Part 2 of 2)}
\label{tab:TrainedfromInstructLowDataScenario_forgetting_part2}
\small
\begin{tabular}{lr}
\toprule
Category & LIMO v2 (32B) \\
\midrule
Common Sense & 3.0 {\scriptsize$\pm$0.3} (84.6) \\
Culture & 4.4 {\scriptsize$\pm$0.9} (62.1) \\
Logic & 6.1 {\scriptsize$\pm$0.4} (80.4) \\
Knowledge/QA & 2.4 {\scriptsize$\pm$0.3} (61.3) \\
Language & 4.2 {\scriptsize$\pm$0.6} (57.0) \\
Liberal Arts & 2.3 {\scriptsize$\pm$0.3} (80.9) \\
Math & 1.7 {\scriptsize$\pm$0.1} (80.0) \\
Safety/Truth & 2.8 {\scriptsize$\pm$0.2} (64.5) \\
Science/Tech & 1.9 {\scriptsize$\pm$0.2} (76.5) \\
\midrule
\textbf{Total} & 3.0 {\scriptsize$\pm$0.2} (74.3) \\
\bottomrule
\end{tabular}
\end{table*}

\begin{table*}[htbp]
\centering
\caption{Trained from Instruct - Low Data Scenario: Backward Transfer (Part 1 of 2)}
\label{tab:TrainedfromInstructLowDataScenario_btransfer_part1}
\small
\begin{tabular}{lrrrr}
\toprule
Category & s1.1 (7B) & s1.1 (14B) & s1.1 (32B) & LIMO (32B) \\
\midrule
Common Sense & 7.3 {\scriptsize$\pm$0.5} (75.8) & 4.6 {\scriptsize$\pm$0.3} (81.5) & 4.5 {\scriptsize$\pm$0.2} (84.3) & 3.9 {\scriptsize$\pm$0.4} (85.1) \\
Culture & 5.7 {\scriptsize$\pm$0.5} (35.0) & 4.6 {\scriptsize$\pm$0.4} (54.2) & 4.6 {\scriptsize$\pm$0.4} (56.7) & 7.5 {\scriptsize$\pm$0.8} (66.1) \\
Logic & 13.8 {\scriptsize$\pm$0.6} (68.8) & 9.2 {\scriptsize$\pm$0.5} (81.1) & 5.7 {\scriptsize$\pm$0.4} (83.9) & 5.8 {\scriptsize$\pm$0.3} (84.6) \\
Knowledge/QA & 13.1 {\scriptsize$\pm$0.3} (59.3) & 11.9 {\scriptsize$\pm$1.3} (69.0) & 11.3 {\scriptsize$\pm$0.5} (71.2) & 11.1 {\scriptsize$\pm$1.3} (71.2) \\
Language & 10.2 {\scriptsize$\pm$0.7} (44.7) & 7.4 {\scriptsize$\pm$1.2} (55.7) & 6.6 {\scriptsize$\pm$0.5} (60.4) & 5.6 {\scriptsize$\pm$0.7} (60.3) \\
Liberal Arts & 6.3 {\scriptsize$\pm$0.7} (71.5) & 5.3 {\scriptsize$\pm$0.6} (78.9) & 5.0 {\scriptsize$\pm$0.1} (82.4) & 4.7 {\scriptsize$\pm$0.2} (83.7) \\
Math & 11.5 {\scriptsize$\pm$0.3} (74.2) & 12.0 {\scriptsize$\pm$0.7} (81.9) & 9.4 {\scriptsize$\pm$0.9} (87.8) & 9.6 {\scriptsize$\pm$0.7} (90.7) \\
Safety/Truth & 9.4 {\scriptsize$\pm$0.1} (52.5) & 7.4 {\scriptsize$\pm$0.9} (62.2) & 5.2 {\scriptsize$\pm$0.4} (65.4) & 5.0 {\scriptsize$\pm$0.4} (67.1) \\
Science/Tech & 8.5 {\scriptsize$\pm$0.1} (66.8) & 7.9 {\scriptsize$\pm$0.3} (77.3) & 7.6 {\scriptsize$\pm$0.5} (82.2) & 7.2 {\scriptsize$\pm$0.3} (83.2) \\
\midrule
\textbf{Total} & 9.1 {\scriptsize$\pm$0.2} (63.5) & 7.2 {\scriptsize$\pm$0.3} (73.5) & 6.2 {\scriptsize$\pm$0.2} (76.9) & 6.2 {\scriptsize$\pm$0.2} (78.8) \\
\bottomrule
\end{tabular}
\end{table*}

\begin{table*}[htbp]
\centering
\caption{Trained from Instruct - Low Data Scenario: Backward Transfer (Part 2 of 2)}
\label{tab:TrainedfromInstructLowDataScenario_btransfer_part2}
\small
\begin{tabular}{lr}
\toprule
Category & LIMO v2 (32B) \\
\midrule
Common Sense & 4.2 {\scriptsize$\pm$0.2} (86.3) \\
Culture & 6.7 {\scriptsize$\pm$0.8} (64.6) \\
Logic & 5.3 {\scriptsize$\pm$0.6} (79.6) \\
Knowledge/QA & 8.0 {\scriptsize$\pm$0.7} (67.7) \\
Language & 5.3 {\scriptsize$\pm$0.3} (59.3) \\
Liberal Arts & 4.9 {\scriptsize$\pm$0.2} (84.4) \\
Math & 10.1 {\scriptsize$\pm$0.8} (90.8) \\
Safety/Truth & 4.5 {\scriptsize$\pm$0.7} (66.7) \\
Science/Tech & 7.5 {\scriptsize$\pm$0.2} (84.0) \\
\midrule
\textbf{Total} & 5.8 {\scriptsize$\pm$0.3} (77.9) \\
\bottomrule
\end{tabular}
\end{table*}

\clearpage

\subsection{Qwen2.5 Base and Coder Merge (Relative to Qwen2.5 Base)}

\begin{table*}[htbp]
\centering
\caption{Qwen2.5 Base and Coder Merge: Forgetting}
\label{tab:FewShotCoderMergeBefore_forgetting_part1}
\small
\begin{tabular}{lrrrr}
\toprule
Category & Linear (0.2) & Linear (0.8) & Slerp (0.2) & Slerp (0.8) \\
\midrule
Common Sense & 8.2 {\scriptsize$\pm$0.6} (67.2) & 11.3 {\scriptsize$\pm$0.7} (67.2) & 13.9 {\scriptsize$\pm$2.2} (67.2) & 16.2 {\scriptsize$\pm$6.4} (67.2) \\
Culture & 12.0 {\scriptsize$\pm$2.0} (60.8) & 15.8 {\scriptsize$\pm$1.8} (60.8) & 21.1 {\scriptsize$\pm$3.4} (60.8) & 41.3 {\scriptsize$\pm$3.3} (60.8) \\
Logic & 11.6 {\scriptsize$\pm$0.4} (51.9) & 19.4 {\scriptsize$\pm$1.0} (51.9) & 23.4 {\scriptsize$\pm$1.1} (51.9) & 17.7 {\scriptsize$\pm$0.9} (51.9) \\
Knowledge/QA & 4.6 {\scriptsize$\pm$0.4} (56.4) & 8.1 {\scriptsize$\pm$0.3} (56.4) & 10.9 {\scriptsize$\pm$1.0} (56.4) & 16.2 {\scriptsize$\pm$1.0} (56.4) \\
Language & 7.8 {\scriptsize$\pm$0.4} (43.2) & 11.2 {\scriptsize$\pm$0.2} (43.2) & 14.0 {\scriptsize$\pm$0.5} (43.2) & 12.7 {\scriptsize$\pm$0.4} (43.2) \\
Liberal Arts & 2.5 {\scriptsize$\pm$0.5} (67.6) & 4.1 {\scriptsize$\pm$0.9} (67.6) & 5.8 {\scriptsize$\pm$1.6} (67.6) & 8.9 {\scriptsize$\pm$0.5} (67.6) \\
Math & 4.5 {\scriptsize$\pm$1.2} (44.9) & 8.9 {\scriptsize$\pm$2.1} (44.9) & 12.4 {\scriptsize$\pm$1.9} (44.9) & 9.4 {\scriptsize$\pm$0.7} (44.9) \\
Safety/Truth & 1.9 {\scriptsize$\pm$0.7} (49.9) & 3.7 {\scriptsize$\pm$0.6} (49.9) & 6.0 {\scriptsize$\pm$0.3} (49.9) & 4.7 {\scriptsize$\pm$1.0} (60.4) \\
Science/Tech & 2.6 {\scriptsize$\pm$0.6} (52.0) & 5.2 {\scriptsize$\pm$0.7} (52.0) & 6.7 {\scriptsize$\pm$1.3} (52.0) & 4.4 {\scriptsize$\pm$0.9} (52.0) \\
\midrule
\textbf{Total} & 6.1 {\scriptsize$\pm$0.2} (58.0) & 9.3 {\scriptsize$\pm$0.4} (58.0) & 12.3 {\scriptsize$\pm$0.2} (58.0) & 13.9 {\scriptsize$\pm$0.4} (59.0) \\
\bottomrule
\end{tabular}
\end{table*}

\begin{table*}[htbp]
\centering
\caption{Qwen2.5 Base and Coder Merge: Backward Transfer}
\label{tab:FewShotCoderMergeBefore_btransfer_part1}
\small
\begin{tabular}{lrrrr}
\toprule
Category & Linear (0.2) & Linear (0.8) & Slerp (0.2) & Slerp (0.8) \\
\midrule
Common Sense & 8.3 {\scriptsize$\pm$0.4} (67.4) & 9.8 {\scriptsize$\pm$0.5} (65.2) & 9.2 {\scriptsize$\pm$0.2} (61.0) & 7.2 {\scriptsize$\pm$0.9} (55.3) \\
Culture & 4.1 {\scriptsize$\pm$0.8} (46.8) & 6.2 {\scriptsize$\pm$1.8} (46.1) & 8.3 {\scriptsize$\pm$2.0} (44.0) & 2.6 {\scriptsize$\pm$0.5} (15.1) \\
Logic & 5.3 {\scriptsize$\pm$0.6} (43.7) & 4.2 {\scriptsize$\pm$0.1} (31.0) & 3.7 {\scriptsize$\pm$0.1} (24.6) & 4.4 {\scriptsize$\pm$0.2} (33.3) \\
Knowledge/QA & 8.3 {\scriptsize$\pm$0.4} (61.6) & 9.2 {\scriptsize$\pm$0.4} (58.2) & 8.2 {\scriptsize$\pm$0.8} (53.4) & 5.1 {\scriptsize$\pm$0.6} (44.1) \\
Language & 4.2 {\scriptsize$\pm$0.8} (39.1) & 7.2 {\scriptsize$\pm$0.4} (37.4) & 6.3 {\scriptsize$\pm$0.2} (32.1) & 4.6 {\scriptsize$\pm$0.5} (30.7) \\
Liberal Arts & 2.1 {\scriptsize$\pm$0.4} (67.3) & 6.1 {\scriptsize$\pm$1.6} (70.2) & 5.7 {\scriptsize$\pm$2.0} (67.6) & 2.1 {\scriptsize$\pm$0.3} (58.7) \\
Math & 5.5 {\scriptsize$\pm$1.0} (45.0) & 6.5 {\scriptsize$\pm$1.6} (39.9) & 5.8 {\scriptsize$\pm$2.9} (34.0) & 4.6 {\scriptsize$\pm$1.4} (36.3) \\
Safety/Truth & 2.4 {\scriptsize$\pm$0.7} (51.8) & 5.0 {\scriptsize$\pm$1.9} (52.7) & 6.8 {\scriptsize$\pm$3.0} (51.5) & 3.0 {\scriptsize$\pm$1.0} (57.9) \\
Science/Tech & 2.7 {\scriptsize$\pm$0.3} (51.4) & 5.6 {\scriptsize$\pm$1.5} (52.3) & 5.2 {\scriptsize$\pm$1.6} (49.7) & 2.3 {\scriptsize$\pm$0.3} (48.1) \\
\midrule
\textbf{Total} & 4.8 {\scriptsize$\pm$0.1} (56.0) & 6.5 {\scriptsize$\pm$0.7} (53.9) & 6.4 {\scriptsize$\pm$0.9} (49.9) & 4.0 {\scriptsize$\pm$0.1} (46.1) \\
\bottomrule
\end{tabular}
\end{table*}

\clearpage

\clearpage

\subsection{Qwen2.5 Base and Coder Merge (Relative to Qwen2.5 Coder)}

\begin{table*}[htbp]
\centering
\caption{Qwen2.5 Base and Coder Merge: Forgetting}
\label{tab:FewShotCoderMergeAfter_forgetting_part1}
\small
\begin{tabular}{lrrrr}
\toprule
Category & Linear (0.2) & Linear (0.8) & Slerp (0.2) & Slerp (0.8) \\
\midrule
Common Sense & 10.2 {\scriptsize$\pm$0.7} (70.5) & 12.7 {\scriptsize$\pm$0.8} (70.5) & 15.2 {\scriptsize$\pm$2.5} (70.5) & 18.2 {\scriptsize$\pm$7.0} (70.5) \\
Culture & 6.8 {\scriptsize$\pm$1.9} (37.8) & 10.6 {\scriptsize$\pm$1.4} (37.8) & 15.8 {\scriptsize$\pm$2.3} (37.8) & 28.5 {\scriptsize$\pm$3.2} (37.8) \\
Logic & 15.1 {\scriptsize$\pm$0.9} (47.2) & 20.4 {\scriptsize$\pm$0.7} (47.2) & 23.2 {\scriptsize$\pm$0.5} (47.2) & 18.8 {\scriptsize$\pm$0.3} (47.2) \\
Knowledge/QA & 6.5 {\scriptsize$\pm$0.8} (52.5) & 8.0 {\scriptsize$\pm$0.6} (52.5) & 9.8 {\scriptsize$\pm$0.3} (52.5) & 14.0 {\scriptsize$\pm$1.9} (52.5) \\
Language & 6.6 {\scriptsize$\pm$0.6} (29.4) & 8.1 {\scriptsize$\pm$0.5} (29.4) & 9.5 {\scriptsize$\pm$1.1} (29.4) & 9.2 {\scriptsize$\pm$0.9} (29.4) \\
Liberal Arts & 6.6 {\scriptsize$\pm$0.1} (63.3) & 6.1 {\scriptsize$\pm$1.0} (63.3) & 7.6 {\scriptsize$\pm$1.7} (63.3) & 12.3 {\scriptsize$\pm$0.3} (63.3) \\
Math & 17.1 {\scriptsize$\pm$0.5} (58.9) & 19.6 {\scriptsize$\pm$1.2} (58.9) & 22.2 {\scriptsize$\pm$2.1} (58.9) & 20.8 {\scriptsize$\pm$0.3} (58.9) \\
Safety/Truth & 10.9 {\scriptsize$\pm$1.7} (42.0) & 10.0 {\scriptsize$\pm$1.4} (42.0) & 10.6 {\scriptsize$\pm$1.3} (42.0) & 11.2 {\scriptsize$\pm$1.2} (43.4) \\
Science/Tech & 10.3 {\scriptsize$\pm$0.4} (54.3) & 10.3 {\scriptsize$\pm$0.9} (54.3) & 11.6 {\scriptsize$\pm$1.9} (54.3) & 11.9 {\scriptsize$\pm$0.3} (54.3) \\
\midrule
\textbf{Total} & 9.5 {\scriptsize$\pm$0.3} (54.0) & 11.1 {\scriptsize$\pm$0.4} (54.0) & 13.4 {\scriptsize$\pm$0.3} (54.0) & 15.3 {\scriptsize$\pm$0.9} (54.1) \\
\bottomrule
\end{tabular}
\end{table*}

\begin{table*}[htbp]
\centering
\caption{Qwen2.5 Base and Coder Merge: Backward Transfer}
\label{tab:FewShotCoderMergeAfter_btransfer_part1}
\small
\begin{tabular}{lrrrr}
\toprule
Category & Linear (0.2) & Linear (0.8) & Slerp (0.2) & Slerp (0.8) \\
\midrule
Common Sense & 7.9 {\scriptsize$\pm$0.4} (67.4) & 8.7 {\scriptsize$\pm$0.5} (65.2) & 8.0 {\scriptsize$\pm$1.3} (61.0) & 6.8 {\scriptsize$\pm$1.6} (55.3) \\
Culture & 13.4 {\scriptsize$\pm$1.7} (46.8) & 15.8 {\scriptsize$\pm$3.2} (46.1) & 17.8 {\scriptsize$\pm$3.6} (44.0) & 6.2 {\scriptsize$\pm$1.9} (15.1) \\
Logic & 11.4 {\scriptsize$\pm$0.2} (43.7) & 7.5 {\scriptsize$\pm$0.3} (31.0) & 6.0 {\scriptsize$\pm$0.2} (24.6) & 7.8 {\scriptsize$\pm$0.0} (33.3) \\
Knowledge/QA & 15.6 {\scriptsize$\pm$1.0} (61.6) & 14.3 {\scriptsize$\pm$0.2} (58.2) & 12.2 {\scriptsize$\pm$1.2} (53.4) & 8.7 {\scriptsize$\pm$1.5} (44.1) \\
Language & 11.0 {\scriptsize$\pm$0.5} (39.1) & 12.2 {\scriptsize$\pm$0.3} (37.4) & 9.9 {\scriptsize$\pm$0.6} (32.1) & 9.1 {\scriptsize$\pm$0.2} (30.7) \\
Liberal Arts & 9.6 {\scriptsize$\pm$0.4} (67.3) & 11.2 {\scriptsize$\pm$0.9} (70.2) & 10.8 {\scriptsize$\pm$1.3} (67.6) & 8.5 {\scriptsize$\pm$0.2} (58.7) \\
Math & 7.7 {\scriptsize$\pm$1.0} (45.0) & 7.0 {\scriptsize$\pm$0.9} (39.9) & 5.1 {\scriptsize$\pm$1.5} (34.0) & 5.2 {\scriptsize$\pm$0.8} (36.3) \\
Safety/Truth & 8.7 {\scriptsize$\pm$1.7} (39.1) & 10.3 {\scriptsize$\pm$1.3} (42.4) & 9.1 {\scriptsize$\pm$1.7} (39.9) & 9.2 {\scriptsize$\pm$0.4} (41.0) \\
Science/Tech & 8.1 {\scriptsize$\pm$0.2} (51.4) & 8.8 {\scriptsize$\pm$0.6} (52.3) & 8.1 {\scriptsize$\pm$0.6} (49.7) & 7.2 {\scriptsize$\pm$0.2} (48.1) \\
\midrule
\textbf{Total} & 10.0 {\scriptsize$\pm$0.3} (54.8) & 10.2 {\scriptsize$\pm$0.6} (52.9) & 9.3 {\scriptsize$\pm$0.7} (48.8) & 7.5 {\scriptsize$\pm$0.1} (44.4) \\
\bottomrule
\end{tabular}
\end{table*}

\clearpage

\subsection{Qwen2.5 Instruct and OpenThinker 7B Merge (Relative to Qwen2.5 Instruct)}

\begin{table*}[htbp]
\centering
\caption{Qwen2.5 Instruct and OpenThinker 7B Merge: Forgetting}
\label{tab:OpenThinker7BMergeBefore_forgetting_part1}
\small
\begin{tabular}{lrr}
\toprule
Category & Linear (0.2) & Linear (0.8) \\
\midrule
Logic & 8.7 {\scriptsize$\pm$1.5} (75.2) & 3.6 {\scriptsize$\pm$0.4} (75.2) \\
Knowledge/QA & 6.5 {\scriptsize$\pm$0.4} (60.3) & 4.2 {\scriptsize$\pm$2.8} (60.3) \\
Liberal Arts & 9.4 {\scriptsize$\pm$0.8} (73.0) & 3.7 {\scriptsize$\pm$0.1} (73.0) \\
Math & 10.5 {\scriptsize$\pm$0.6} (65.3) & 3.3 {\scriptsize$\pm$0.5} (65.3) \\
Safety/Truth & 7.7 {\scriptsize$\pm$1.3} (55.1) & 2.8 {\scriptsize$\pm$0.4} (55.1) \\
Science/Tech & 9.6 {\scriptsize$\pm$0.3} (67.4) & 4.2 {\scriptsize$\pm$0.2} (67.4) \\
\midrule
\textbf{Total} & 8.7 {\scriptsize$\pm$0.2} (66.1) & 3.6 {\scriptsize$\pm$0.5} (66.1) \\
\bottomrule
\end{tabular}
\end{table*}

\begin{table*}[htbp]
\centering
\caption{Qwen2.5 Instruct and OpenThinker 7B Merge: Backward Transfer}
\label{tab:OpenThinker7BMergeBefore_btransfer_part1}
\small
\begin{tabular}{lrr}
\toprule
Category & Linear (0.2) & Linear (0.8) \\
\midrule
Logic & 6.0 {\scriptsize$\pm$0.7} (71.6) & 4.6 {\scriptsize$\pm$1.3} (76.6) \\
Knowledge/QA & 5.2 {\scriptsize$\pm$1.8} (58.6) & 3.8 {\scriptsize$\pm$0.9} (59.8) \\
Liberal Arts & 4.6 {\scriptsize$\pm$0.5} (66.6) & 3.9 {\scriptsize$\pm$0.6} (73.3) \\
Math & 8.8 {\scriptsize$\pm$0.3} (63.1) & 6.8 {\scriptsize$\pm$0.5} (70.1) \\
Safety/Truth & 5.8 {\scriptsize$\pm$0.3} (52.7) & 3.4 {\scriptsize$\pm$0.6} (56.1) \\
Science/Tech & 6.3 {\scriptsize$\pm$0.3} (62.9) & 4.9 {\scriptsize$\pm$0.3} (68.1) \\
\midrule
\textbf{Total} & 6.1 {\scriptsize$\pm$0.1} (62.6) & 4.6 {\scriptsize$\pm$0.2} (67.3) \\
\bottomrule
\end{tabular}
\end{table*}

\clearpage

\subsection{Qwen2.5 Instruct and OpenThinker 7B Merge (Relative to OpenThinker)}

\begin{table*}[htbp]
\centering
\caption{Qwen2.5 Instruct and OpenThinker 7B Merge: Forgetting}
\label{tab:OpenThinker7BMergeAfter_forgetting_part1}
\small
\begin{tabular}{lrr}
\toprule
Category & Linear (0.2) & Linear (0.8) \\
\midrule
Logic & 7.1 {\scriptsize$\pm$0.8} (64.0) & 3.7 {\scriptsize$\pm$1.7} (64.0) \\
Knowledge/QA & 7.3 {\scriptsize$\pm$1.6} (52.2) & 7.0 {\scriptsize$\pm$1.2} (52.2) \\
Liberal Arts & 8.5 {\scriptsize$\pm$0.2} (59.2) & 5.5 {\scriptsize$\pm$0.4} (59.2) \\
Math & 14.3 {\scriptsize$\pm$1.6} (72.1) & 9.4 {\scriptsize$\pm$1.5} (72.1) \\
Safety/Truth & 10.3 {\scriptsize$\pm$2.3} (52.2) & 7.2 {\scriptsize$\pm$1.3} (52.2) \\
Science/Tech & 9.2 {\scriptsize$\pm$0.2} (60.2) & 6.8 {\scriptsize$\pm$0.6} (60.2) \\
\midrule
\textbf{Total} & 9.4 {\scriptsize$\pm$0.6} (60.0) & 6.6 {\scriptsize$\pm$0.6} (60.0) \\
\bottomrule
\end{tabular}
\end{table*}

\begin{table*}[htbp]
\centering
\caption{Qwen2.5 Instruct and OpenThinker 7B Merge: Backward Transfer}
\label{tab:OpenThinker7BMergeAfter_btransfer_part1}
\small
\begin{tabular}{lrr}
\toprule
Category & Linear (0.2) & Linear (0.8) \\
\midrule
Logic & 12.8 {\scriptsize$\pm$1.6} (71.6) & 13.1 {\scriptsize$\pm$3.7} (76.6) \\
Knowledge/QA & 12.1 {\scriptsize$\pm$2.5} (58.6) & 12.6 {\scriptsize$\pm$2.0} (59.8) \\
Liberal Arts & 14.1 {\scriptsize$\pm$0.2} (66.6) & 16.1 {\scriptsize$\pm$0.6} (73.3) \\
Math & 7.5 {\scriptsize$\pm$1.5} (63.1) & 7.9 {\scriptsize$\pm$1.1} (70.1) \\
Safety/Truth & 10.7 {\scriptsize$\pm$1.0} (52.7) & 10.1 {\scriptsize$\pm$1.1} (56.1) \\
Science/Tech & 11.2 {\scriptsize$\pm$0.8} (62.9) & 12.7 {\scriptsize$\pm$0.4} (68.1) \\
\midrule
\textbf{Total} & 11.4 {\scriptsize$\pm$0.7} (62.6) & 12.1 {\scriptsize$\pm$1.0} (67.3) \\
\bottomrule
\end{tabular}
\end{table*}

\clearpage

\subsection{Qwen2.5 Instruct and OpenThinker3 7B Merge (Relative to Qwen2.5 Instruct)}

\begin{table*}[htbp]
\centering
\caption{Qwen2.5 Instruct and OpenThinker3 7B Merge: Forgetting}
\label{tab:OpenThinker37BMergeBefore_forgetting_part1}
\small
\begin{tabular}{lrr}
\toprule
Category & Linear (0.2) & Linear (0.8) \\
\midrule
Logic & 33.6 {\scriptsize$\pm$2.7} (75.2) & 28.1 {\scriptsize$\pm$1.8} (74.2) \\
Knowledge/QA & 25.7 {\scriptsize$\pm$8.0} (60.3) & 18.0 {\scriptsize$\pm$5.2} (45.0) \\
Liberal Arts & 35.0 {\scriptsize$\pm$2.9} (74.4) & 28.8 {\scriptsize$\pm$1.7} (73.8) \\
Math & 40.1 {\scriptsize$\pm$6.1} (65.3) & 30.0 {\scriptsize$\pm$0.7} (65.3) \\
Safety/Truth & 36.2 {\scriptsize$\pm$2.8} (64.1) & 22.8 {\scriptsize$\pm$3.7} (61.0) \\
Science/Tech & 30.1 {\scriptsize$\pm$3.2} (67.4) & 29.6 {\scriptsize$\pm$0.9} (67.4) \\
\midrule
\textbf{Total} & 33.5 {\scriptsize$\pm$3.7} (67.8) & 26.0 {\scriptsize$\pm$0.8} (63.8) \\
\bottomrule
\end{tabular}
\end{table*}

\begin{table*}[htbp]
\centering
\caption{Qwen2.5 Instruct and OpenThinker3 7B Merge: Backward Transfer}
\label{tab:OpenThinker37BMergeBefore_btransfer_part1}
\small
\begin{tabular}{lrr}
\toprule
Category & Linear (0.2) & Linear (0.8) \\
\midrule
Logic & 1.3 {\scriptsize$\pm$2.0} (31.3) & 2.3 {\scriptsize$\pm$1.0} (39.9) \\
Knowledge/QA & 1.9 {\scriptsize$\pm$1.0} (28.8) & 1.2 {\scriptsize$\pm$1.4} (21.7) \\
Liberal Arts & 1.9 {\scriptsize$\pm$0.8} (30.2) & 2.9 {\scriptsize$\pm$0.1} (39.2) \\
Math & 0.7 {\scriptsize$\pm$0.8} (12.6) & 1.6 {\scriptsize$\pm$0.8} (27.0) \\
Safety/Truth & 0.9 {\scriptsize$\pm$1.1} (15.8) & 2.9 {\scriptsize$\pm$0.6} (34.6) \\
Science/Tech & 2.5 {\scriptsize$\pm$1.0} (30.4) & 2.5 {\scriptsize$\pm$0.2} (30.9) \\
\midrule
\textbf{Total} & 1.5 {\scriptsize$\pm$1.1} (24.9) & 2.2 {\scriptsize$\pm$0.3} (31.7) \\
\bottomrule
\end{tabular}
\end{table*}

\clearpage

\subsection{Qwen2.5 Instruct and OpenThinker3 7B Merge (Relative to OpenThinker3)}

\begin{table*}[htbp]
\centering
\caption{Qwen2.5 Instruct and OpenThinker3 7B Merge: Forgetting}
\label{tab:OpenThinker37BMergeAfter_forgetting_part1}
\small
\begin{tabular}{lrr}
\toprule
Category & Linear (0.2) & Linear (0.8) \\
\midrule
Logic & 28.1 {\scriptsize$\pm$3.2} (65.4) & 25.1 {\scriptsize$\pm$2.6} (66.9) \\
Knowledge/QA & 21.2 {\scriptsize$\pm$6.4} (54.3) & 17.2 {\scriptsize$\pm$7.3} (38.6) \\
Liberal Arts & 31.3 {\scriptsize$\pm$2.9} (68.9) & 27.2 {\scriptsize$\pm$1.7} (68.3) \\
Math & 52.3 {\scriptsize$\pm$6.9} (81.1) & 41.8 {\scriptsize$\pm$1.8} (81.1) \\
Safety/Truth & 33.5 {\scriptsize$\pm$2.1} (59.8) & 23.5 {\scriptsize$\pm$1.7} (58.6) \\
Science/Tech & 32.0 {\scriptsize$\pm$3.5} (70.7) & 32.2 {\scriptsize$\pm$1.4} (70.5) \\
\midrule
\textbf{Total} & 33.1 {\scriptsize$\pm$3.8} (66.7) & 28.0 {\scriptsize$\pm$1.2} (63.8) \\
\bottomrule
\end{tabular}
\end{table*}

\begin{table*}[htbp]
\centering
\caption{Qwen2.5 Instruct and OpenThinker3 7B Merge: Backward Transfer}
\label{tab:OpenThinker37BMergeAfter_btransfer_part1}
\small
\begin{tabular}{lrr}
\toprule
Category & Linear (0.2) & Linear (0.8) \\
\midrule
Logic & 3.0 {\scriptsize$\pm$5.0} (31.3) & 4.9 {\scriptsize$\pm$0.9} (39.9) \\
Knowledge/QA & 2.2 {\scriptsize$\pm$0.7} (28.8) & 4.5 {\scriptsize$\pm$2.3} (21.7) \\
Liberal Arts & 2.3 {\scriptsize$\pm$0.8} (30.2) & 5.4 {\scriptsize$\pm$0.7} (39.2) \\
Math & 0.5 {\scriptsize$\pm$0.6} (12.6) & 1.5 {\scriptsize$\pm$0.2} (27.0) \\
Safety/Truth & 1.0 {\scriptsize$\pm$0.8} (15.8) & 5.5 {\scriptsize$\pm$0.4} (34.6) \\
Science/Tech & 2.0 {\scriptsize$\pm$0.7} (30.4) & 2.8 {\scriptsize$\pm$0.3} (30.9) \\
\midrule
\textbf{Total} & 1.8 {\scriptsize$\pm$1.3} (24.9) & 4.0 {\scriptsize$\pm$0.4} (31.7) \\
\bottomrule
\end{tabular}
\end{table*}

\clearpage

\section*{Disclaimer for use of LLMs}
We primarily used LLMs in coding co-pilot applications to facilitate experimentation and help with plotting code for result presentation. LLMs were also used as writing tools to assist in refining the paper. However, the final version was carefully reviewed and finalized by the authors. No LLMs were used in ideation and experimental design.

\end{document}